\documentclass{article}

\usepackage{tikz}
\usetikzlibrary{quotes}

\usepackage{xspace}
\usepackage{paralist}

\newenvironment{CompactItemize}{
\begin{list}{$\bullet$}{%
\setlength{\leftmargin}{14pt}
\setlength{\itemindent}{0pt}
}}
{\end{list}}
\usepackage{listings}
\usepackage[export]{adjustbox}
\usepackage{xfrac}
\usepackage{amsmath,amssymb}
\usepackage{amsthm}
\usepackage{mathtools}
\usepackage{fancyvrb}
\usepackage{soul}
\usepackage[noend]{algorithmic}
\usepackage[ruled,vlined]{algorithm2e}
\usepackage{url}
\usepackage{fullpage}
\usepackage{makeidx}
\usepackage{graphicx,float,psfrag,epsfig,subcaption}
\usepackage[font=small]{caption}

\usepackage{hyperref}
\usepackage{epstopdf}
\usepackage{color}
\usepackage{xr}

\usepackage{caption}
\usepackage[utf8]{inputenc}
\usepackage{thm-restate}

\usepackage{scalerel,stackengine}
\usepackage{authblk}

\usepackage{enumitem}

\stackMath
\newcommand\reallywidehat[1]{%
\savestack{\tmpbox}{\stretchto{%
  \scaleto{%
    \scalerel*[\widthof{\ensuremath{#1}}]{\kern.1pt\mathchar"0362\kern.1pt}%
    {\rule{0ex}{\textheight}}
  }{\textheight}%
}{2.4ex}}%
\stackon[-6.9pt]{#1}{\tmpbox}%
}
\usepackage[mathscr]{euscript}

\numberwithin{equation}{section}

\newtheoremstyle{myexample} 
    {\topsep}                    
    {\topsep}                    
    {\rm }                   
    {}                           
    {\bf }                   
    {.}                          
    {.5em}                       
    {}  

\newtheoremstyle{myremark} 
    {\topsep}                    
    {\topsep}                    
    {\rm}                        
    {}                           
    {\bf}                        
    {.}                          
    {.5em}                       
    {}  

\theoremstyle{myremark}

\theoremstyle{myremark}

\theoremstyle{myexample}

\newcommand{\E}{\mathbb E}

\newcommand{\R}{\mathbb R}

\begin{document}

\title{Unveiling Transformers with LEGO: 
a synthetic reasoning task}

\author{Yi Zhang}
\author{Arturs Backurs}
\author{S\'ebastien Bubeck\\}
\author{Ronen Eldan}
\author{Suriya Gunasekar}
\author{Tal Wagner}
\affil{\bf Microsoft Research \authorcr{\small\tt\{zhayi, arturs.backurs, sebubeck, roneneldan, suriya.gunasekar, tal.wagner\}@microsoft.com}}

\date{}

\title{Unveiling Transformers with LEGO: \\
a synthetic reasoning task}

\author{Yi Zhang, Arturs Backurs, S\'ebastien Bubeck, \\
Ronen Eldan, Suriya Gunasekar, Tal Wagner \\
Microsoft Research}

\date{}
\maketitle\begin{abstract}
We propose a synthetic reasoning task, LEGO (Learning Equality and Group Operations), that encapsulates the problem of following a chain of reasoning, and we study how the Transformer architectures learn this task. We pay special attention to data effects such as pretraining (on seemingly unrelated NLP tasks) and dataset composition (e.g., differing chain length at training and test time), as well as architectural variants such as weight-tied layers or adding convolutional components. We study how the trained models eventually succeed at the task, and in particular, we manage to understand some of the attention heads as well as how the information flows in the network. In particular, we have identified a novel \emph{association} pattern that globally attends only to identical tokens. Based on these observations we propose a hypothesis that here pretraining helps for LEGO tasks due to certain structured attention patterns, and we experimentally verify this hypothesis. We also observe that in some data regime the trained transformer finds ``shortcut" solutions to follow the chain of reasoning, which impedes the model's robustness, and moreover we propose ways to prevent it. Motivated by our findings on structured attention patterns, we propose the LEGO attention module, a drop-in replacement for vanilla attention heads. This architectural change significantly reduces Flops and maintains or even \emph{improves} the model's performance at large-scale pretraining.
\end{abstract}

\section{Introduction} \label{sec:intro}
The deep learning revolution is about training large neural networks on vast amount of data. The first field transformed by this methodology was computer vision, crucially leveraging the convolutional neural network architecture \cite{LeCun89, KSH12}. More recently natural language processing was revolutionized by the Transformer architecture \cite{Vas17}. Transformers are designed to process input represented as ``set of elements" (e.g., the words in a sentence with their positional encoding). This is of course an incredibly generic assumption, and thus Transformers can be applied to a wide variety of tasks, including vision \cite{ViT}, reinforcement learning \cite{chen2021decision}, and protein structure prediction \cite{rives2021biological,jumper2021highly} among others, or even jointly across domains to produce generalized agents \cite{reed2022generalist}. In fact, learning with Transformers is rapidly becoming the norm in deep learning.


Transformer models display excellent performance on the standard criterion ``training error/test error" (e.g., for masked language prediction or translation). However, what makes them particularly noteworthy, is that large-scale Transformer models seem to exhibit unexpected emergent behaviors, such as basic reasoning ability \cite{thoppilan2022lamda, GPT3,chowdhery2022palm, du2021glam, rae2021scaling, hoffmann2022training, smith2022using, zhang2022opt, wei2022chain, nye2022show}, excellent fine-tuning performance \cite{hu2022lora, thoppilan2022lamda, nye2022show, rae2021scaling, polu2022formal}, or zero-shot learning \cite{GPT3,chowdhery2022palm, du2021glam, rae2021scaling, hoffmann2022training, smith2022using, zhang2022opt}. 
Currently, there is a remarkable community effort towards {\em at-scale} experimental investigation of Transformers, essentially trying to find out what such models can do when they become large enough and are trained on large/diverse enough datasets. The successes are striking and capture the imagination \cite{GPT3, dalle2}. Yet, for all of these wonders, there is very little {\em understanding} of how these models learn, or in fact what {\em do} they learn. Answering such questions in the {\em at-scale} experiments is particularly challenging, as one has little control over the data when hundreds of billions of tokens are harvested from various sources. In this paper, we propose to take a step back, and try to understand how learning occurs and what is being learned in a more controlled setting that captures important aspects of ``reasoning".

The benefit of such a controlled setting is that we can try to understand some of the most pressing questions in learning with Transformers, particularly around (i) the architecture  and (ii) the importance of training data. For (i) we probe the role of multiple heads and depth, and we show that we can successfully understand them in our controlled setting. For (ii) we investigate how much the dataset composition matters, as well as how pretraining on merely vaguely related tasks makes fine-tuning successful. In turn, these insights can guide our thinking for large-scale experiments, and we give some of the lessons learned below. In particular, our insights crystallize into an architectural change to BERT for faster inference with matching or even better performance  (Section~\ref{sec:LEGO v0}).

\subsection{LEGO: A synthetic reasoning task}
Core components of reasoning include the ability to {\em associate} concepts, and to {\em manipulate} them. We propose a simple task that captures these two aspects, which we call LEGO (Learning Equality and Group Operations). In LEGO, the input describes a sequence of {\em variable assignments} as well as {\em operations} on these variables by a fixed (mathematical) group. One needs to be able to deal with both long-range assignments (the same variable appearing in different parts of the input should be viewed as a being {\em equal} to same quantity), as well as short-range operations (describing what group element is applied to which variable). A key parameter of an input sequence will be its length, which is proportional to the number of sequential reasoning steps one has to do in order to resolve the value of each variable. We will mostly train with a fixed sequence length (say $12$). We often provide supervision only on part of the sequence (say the first $6$ variables). We do so in order to test the generalization capabilities from smaller length sequences to longer length sequences without introducing potential errors due to the positional encoding in Transformers.

\subsection{Some takeaways} \label{sec:takeaways}
In LEGO, we are interested in both {\em classical generalization} (i.e., training and test distribution are the same) and {\em out-of-distribution generalization}. For the latter we focus on distribution shifts that vary the length of the chain of reasoning, and thus we refer to this type of generalization as {\em length extrapolation}. Specifically, the setting for length extrapolation is to train with supervision on shorter sequence lengths (e.g., supervision on only the first $6$ variables) and test on a long sequences (e.g., accuracy computed on $12$ variables).
A summary of our empirical observations is as follows:
\begin{CompactItemize}
\item[1.] First, classical generalization happens reliably for all architectures and data regimes.
\item[2.] More interestingly, length extrapolation seems to depend on architectural/data composition choices. Specifically, BERT-like models without special data preparation do {\em not} extrapolate to longer sequences, while other models like ALBERT, or BERT with carefully selected data (such as diverse sequence lengths, or pre-trained BERT) {\em do} extrapolate.
\item[3.] The extrapolating models all seem to evolve attention heads dedicated to either {\em association} (long-range identity matching) or {\em manipulation} (short-range operations). We provide evidence that pre-trained BERT (which is pre-trained on a seemingly unrelated dataset) generalizes because it has learned such heads.
\item[4.] The non-extrapolating models seem to solve the classical generalization problem using a certain shortcut-like solution, whereby using the specificity of the group operations they are able to jump to the end of the chain of reasoning, and then complete the rest of the variables by following the reasoning both from the start {\em and} the end of the chain.
\end{CompactItemize}

We interpret our findings as follows:
\begin{CompactItemize}
\item[(i)] Classical generalization can be a deceptive metric, as there might be unexpected ways to solve the problem. This is famously related to the issue of embedding machine learning systems with {\em common sense reasoning}. Namely, we hope that when an ML system solves a task, it does so in ``the way humans do it", but of course, nothing guarantees that this will happen. Our findings are consistent with the current methodology of increasing the diversity of the training data, which seems crucial for generalization.
\item[(ii)] ALBERT-like models, where a layer is repeated several times, seem to be an ideal structure for problems that could be described algorithmically as a ``for loop" (as is the case with following a chain of reasoning). Indeed we find that ALBERT extrapolates in data regimes where BERT does not, clearly separating these two architectures.
\item[(iii)] The success of pretraining/fine-tuning in vastly different tasks might actually come from a ``simple" better initialization, rather than complex knowledge encoded during pre-training.
\item[(iv)] The interplay between short-range (close-by information in a sentence) and long-range (the same concept appearing in different places in the sentence) is relevant more broadly than in our synthetic task. We observe that the networks effectively learn to deal with short-range/long-range information by implementing specific attention patterns. This motivates us to study a new LEGO attention architecture, and we show it matches or even outperforms its baseline on the large-scale pretraining but with significantly less computational cost.
\end{CompactItemize}

\subsection{Related works} \label{sec:related}
In \cite{PVR}, the PVR (Pointer Value Retrieval) task is introduced, with a similar high-level goal to ours in introducing the LEGO task, namely to study how neural networks learn to reason in a controlled setting. In a PVR task, part of the input indicates another part of the input where a function of potentially varying complexity has to be computed. Like us, they use distribution shift to investigate how various network architectures learn this task, and they observe that networks can learn the task at hand (``classical generalization") yet fail to extrapolate to mild distribution shift. They then ask the following questions:
``Are there architectural changes that can enforce better priors and withstand distribution shift? Can novel learning objectives prevent these adversarial correlations? Progress on these questions holds promise for greater robustness."

Our study attacks these questions directly in the context of the LEGO task (e.g., ALBERT versus BERT, and training set composition investigations), and our preliminary results indicate that this is indeed a fruitful direction to obtain better models in some aspects (e.g., more interpretable). Other examples of recent synthetic benchmark with a similar philosophy include SCAN (Simplified version of the CommAI Navigation) \cite{lake2018generalization}, CFQ (Compositional Freebase Questions) \cite{keysers2020measuring}, LIME \cite{wu2021lime},  PCFG SET \cite{hupkes2020compositionality}, and BONGARD-LOGO \cite{nie2020bongard}. In SCAN for example, one has to ``translate" a command of the form ``turn left twice and jump" into a sequence of actions ``LTURN LTURN JUMP" (see \cite{patel2022revisiting} for more recent progress on this dataset). Again, similarly to the PVR tasks, these works focus on understanding generalization (in these cases, {\em compositional generalization}). Another related line of works is on studying Transformers to recognize various formal languages, see e.g., \cite{bhattamishra-etal-2020-ability, yao2021self}. A contemporary work~\cite{csordas2021neural} proposed modifications to Transformer architectures to achieve significantly better length extrapolation (other works studying this important class of distribution shifts include \cite{anil2022exploring}). As far as we know, none of these works try to probe the inner workings of the networks in the same depth as we do here. On the other hand, networks trained on real data are being extensively scrutinized, see for example \cite{rogers2020primer} where they try to understand some of the attention heads of BERT (see also \cite{saha2020prover}). However, making sense of these real-data-trained networks is a daunting task, and a key contribution of ours is to show in a limited setting one can obtain a clearer picture of what Transformers learn.

The LEGO task is also naturally related to the growing literature on testing mathematical/coding abilities of Transformers (e.g., \cite{DBLP:conf/emnlp/SahaGSB20}), specifically the simpler tasks of checking the correctness of a proof (or simplifying one, such as in \cite{agarwal2021analyzing} which studies simplification of polynomials), or executing code for a given input \cite{chen2021latent}. It would be interesting to see if some of the insights we derive in the present paper apply to currently challenging mathematical tasks such as MATH \cite{hendrycks2021measuring} and IsarStep \cite{li2021isarstep}.

There are an abundance of studies on attention heads that have identified the importance of local, convolutional, attention patterns~\cite{voita2019analyzing, Correia2019AdaptivelyST, clark-etal-2019-bert, raganato2020fixed, you2020hard}. However, to the best of our knowledge, we are the first to demonstrate the importance of the association pattern that globally attends to identical tokens, thanks to the LEGO task.

\section{Learning equality and group operations (LEGO)} \label{sec:LEGO}
We propose the following synthetic task, which we call LEGO. Let $G$ be a finite (semi)group acting on a finite set $X$, and denote $g(x)$ for the action of $g \in G$ on $x \in X$. We define a formal language using the symbols from $G$ and $X$ as well as symbols from a finite alphabet $A$ which we refer to as the {\em variables}. A sentence in our formal language is made of clauses separated by a semi-colon. A clause is of the form $a = g x$ with $a \in A$, $g \in G$ and either $x \in X$ or $x \in A$. If $x \in X$, such a clause means that the variable $a$ is assigned the element $g(x) \in X$. On the other hand if $x \in A$ and the variable $x$ was assigned an element $y \in X$ through another clause (or chain of clauses) in the sentence, then the clause $a = g x$ assigns variable $a$  to the element $g(y) \in X$. The task's goal is to take in input a sentence with a fixed number $n$ of clauses, given in an arbitrary order, and to output the assigned element to each variable that appear in the sentence (the formal language will have a further restriction that ensures that each variable is assigned one and only one element).

\begin{figure}[t]
\vspace{-0.5cm}
\begin{center}
\begin{tikzpicture}
\tikzset{vertex/.style = {shape=circle,draw,minimum size=1.5em}}
\tikzset{edge/.style = {->,> = latex}}
\node[vertex] (1) at  (0,0) {$1$};
\node[vertex] (2) at  (2,0) {$a$};
\node[vertex] (3) at  (4,0) {$b$};
\node[vertex] (4) at  (6,0) {$e$};
\node[vertex] (5) at  (8,0) {$f$};
\node[vertex] (6) at  (10,0) {$d$};
\node[vertex] (7) at  (12,0) {$c$};
\draw[edge]  (1) to["$+$"] (2);
\draw[edge]  (2) to["$-$"] (3);
\draw[edge]  (3) to["$+$"] (4);
\draw[edge]  (4) to["$+$"] (5);
\draw[edge]  (5) to["$-$"] (6);
\draw[edge]  (6) to["$+$"] (7);

\end{tikzpicture}
\end{center}
\vspace{-0.2cm}
\caption{The graph representation of the sentence $a=+1; \, b=-a; \, e=+b; \, d=-f; \, c=+d; \, f=+e$}
\label{fig:example}
\vspace{-0.5cm}
\end{figure}
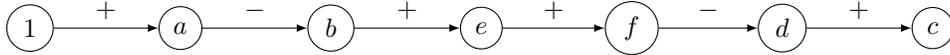

We can view a sentence as a directed graph on the vertex set $X \cup A$ with labelled edges as follows: a clause $a = g x$ corresponds to a directed edge from the vertex $x$ to the vertex $a$, and the edge is labelled with $g$. We restrict our attention to sentences corresponding to a line graph directed away from some fixed root vertex $r \in X$, and whose non-root vertices are all in $A$, see Figure~\ref{fig:example} for an example. In particular such sentences are ``consistent", meaning that a variable is assigned a unique element (the assignment is obtained by simply ``following the chain").

\textbf{Task 1.} The most basic instantiation of LEGO is when $G$ is the unique group of $2$ elements acting on a set $X$ also of $2$ elements, that is $G = \{+, -\}$ and $X = \{1,-1\}$. 
Our sentences thus consists of $n$ clauses of the form $a_i=\pm a_{i-1}$, where $a_i\in A$ for $i=1,2,\ldots,n$ and $a_0=1$ (we fix $r=1$). 
Note that in this case our formal language has well over a billion unique valid sentences when $n \geq 10$. Example of a sentence with $n=6$ is  (see Figure \ref{fig:example} for the graph depiction): $
a=+1; \, b=-a; \, e=+b; \, d=-f; \, c=+d; \, f=+e$.
Our task's goal is to report the elements or values from $X$ assigned to the variables appearing in the sentence. In the above example, assignments for variables  $a,b,c,d,e,f$ are $1,-1,-1,-1,1,1$. 

\textbf{Task 2.} One can think of Task 1 as the case of LEGO for the permutation group on $N=2$ elements (acting on itself). Our second task will correspond to $N=3$, which is qualitatively different since the permutation group on $3$ elements is non-abelian. 

We will focus on Task 1 in the main paper and include in Appendix~\ref{sec:dihedral} experiments on this Task 2. Our training and test data for the task consists of $n$ length chains as described above with the order of clauses in the sentence randomized. A sample input sentence to a transformer looks like \texttt{[BOS] j=-f; f=-b; y=+t; o=+e; d=+y; v=+d; h=-o; b=-i; i=+1; t=+l; e=-j; l=-h; [EOS]}.  See appendix for further data generation details. 

\section{Transformers for LEGO}\label{sec:transformers}
\begin{figure}[h]
    \centering
    \includegraphics[width=15cm]{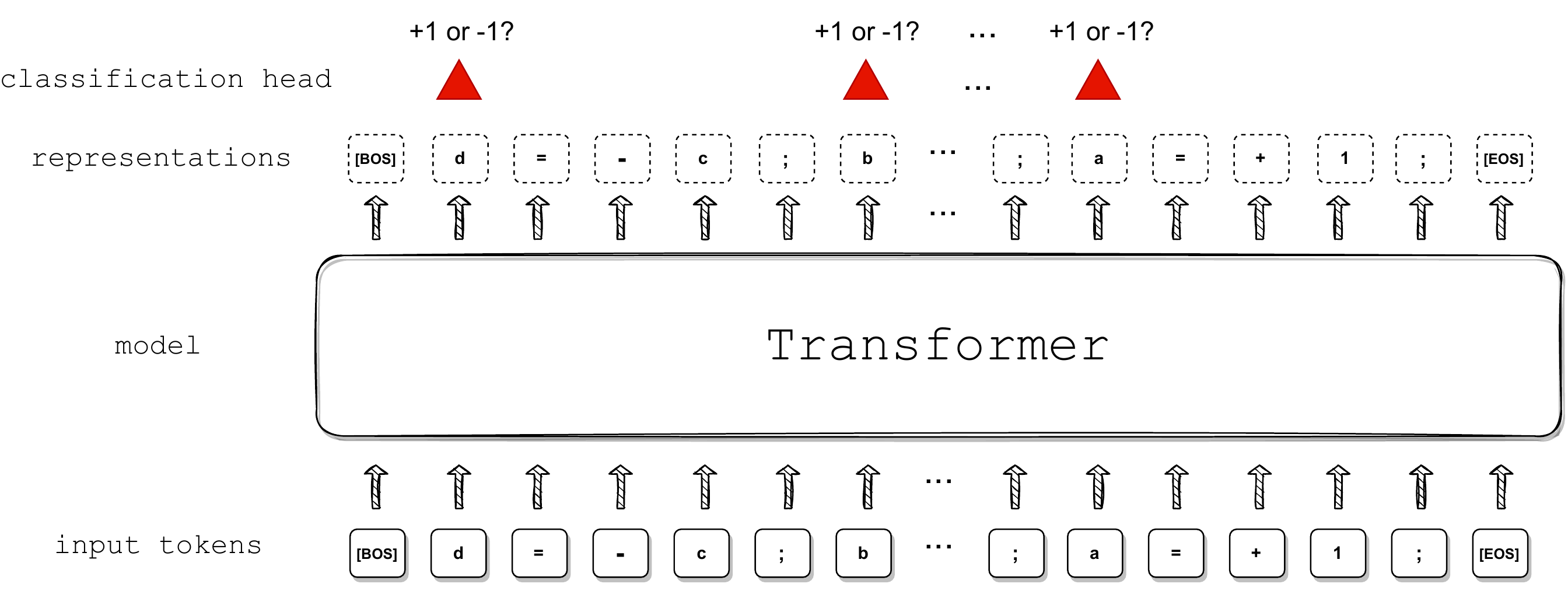}
    \caption{Illustration of a transformer model applied to LEGO Task 1 on input sentence \texttt{d=-c;\,b=-a;\, c=+b;\, a=+1;}. 
    We apply a linear classification head to the output representations of each clause's first token to generate predictions for the variables assignment. }
    \label{fig:transformer-lego-diagram}
\end{figure}


We apply transformer models in the token classification pipeline to predict the assignments of the variables in the input sentence, depicted in Figure~\ref{fig:transformer-lego-diagram}. To evaluate the out-of-distribution generalization (referred to simply as generalization), we introduce the notation of $n_{tr}\le n$, such that during training, supervision is provided only on the first $n_{tr}$ clauses (first in the graph representation of the input sentence). 
We mainly focus on BERT \cite{devlin2018bert} and ALBERT \cite{lan2019albert} architectures. These two models are representative large transformer architectures for NLP tasks, and we observe they exhibit intriguing behavior difference on our tasks which we will detail in Section~\ref{sec:4}. See appendix for training hyper-parameters and dataset construction details.

\begin{figure}[t]
    \centering
    \begin{subfigure}{\textwidth}
     \begin{minipage}[r]{0.05\textwidth}
     \caption{}
     \label{fig:shortcut}
  \end{minipage}
  \begin{minipage}[l]{0.95\textwidth}
    \includegraphics[trim={0cm 0cm 0cm 0cm},clip, width=13cm]{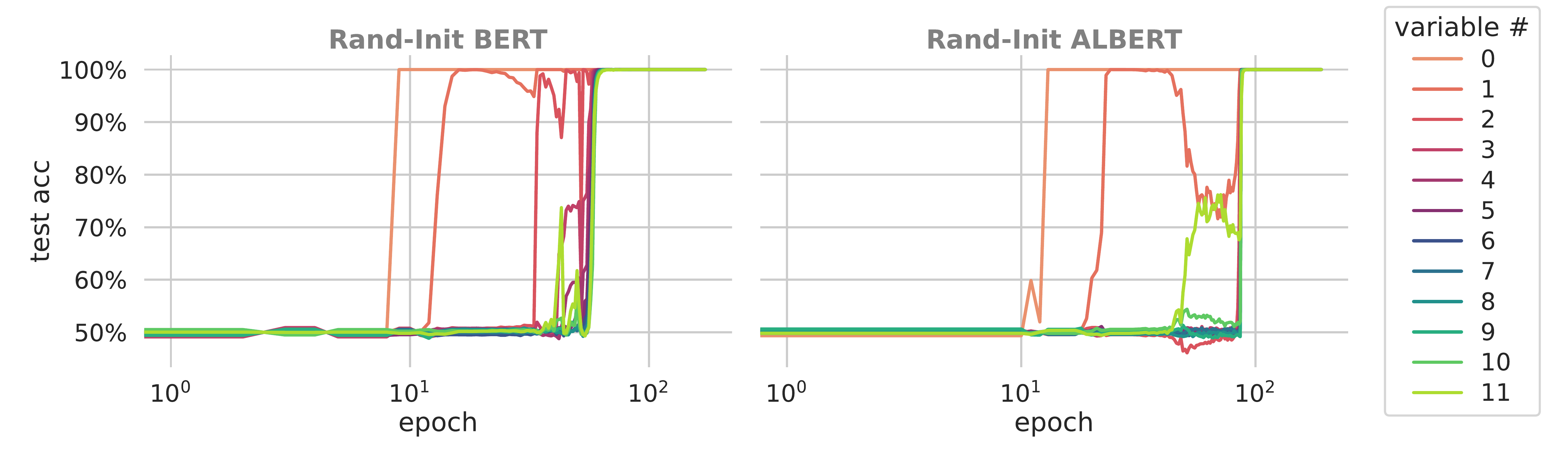}
    \end{minipage}
    \end{subfigure}
    \hrule
\begin{subfigure}{\textwidth}
     \begin{minipage}[r]{0.05\textwidth}
     \caption{}
  \end{minipage}
  \begin{minipage}[l]{0.95\textwidth}
    \includegraphics[trim={0cm .5cm 0cm 0cm},clip, width=15cm]{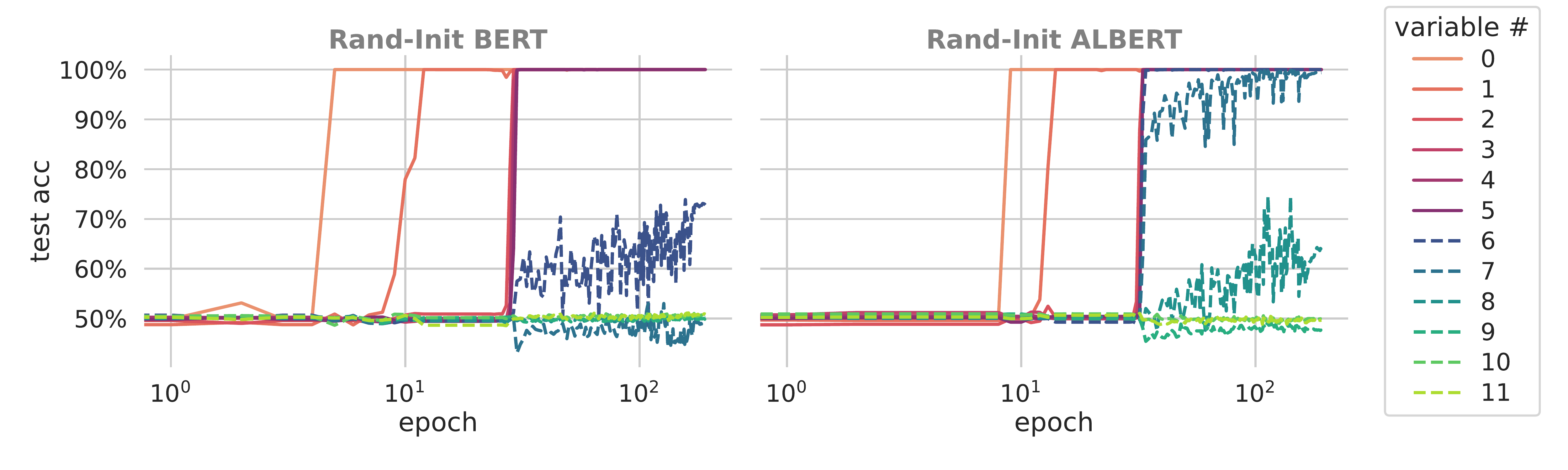}
    \end{minipage}
    \end{subfigure}
    \caption{Solving LEGO (Task 1) using BERT  and ALBERT, trained from random initialization. Each curve corresponds to the test accuracy of a single variable appearing in the sentence over the course of training. The variable numbers in the legend are their position in the reasoning chain (or graph representation) of the input sentence, rather than the position in the sentence itself. For example, on the input sentence: \texttt{b=-a;\,d=-c;\,c=+b;\,a=+1;}, variable $\# 0$ is \texttt{a}, $\#1$ is \texttt{b}, $\#2$ is \texttt{c}, and $\#3$ is \texttt{d}. Top a): models are trained to fit all variables ($n=12, n_{tr}=12$). Bottom b): models are trained to fit the first $6$ variables but test on all 12 variables ($n=12, n_{tr}=6$). Dashed curves represent variables not supervised during training.}
    \label{fig:result_shortcut}
\end{figure}

In Figure~\ref{fig:result_shortcut}, we report initial results on LEGO with $n=12$ and $n_{tr}=6, 12$. 
Both BERT and ALBERT are able to achieve good classical generalization, while only ALBERT appears to generalize even to slightly longer sequence length. We observe similar behavior across different lengths of inputs too. This suggests that classical generalization might be a deceptive metric to evaluate learning of true logic/reasoning tasks. 
Motivated by these initial results, in the next section we focus on breaking down the learning dynamics of BERT and ALBERT for the LEGO task towards carefully understanding their strengths and weaknesses.

\section{Unveiling Transformers with LEGO}
\label{sec:4}

\subsection{BERT vs. ALBERT: Iterative reasoning in iterative architectures}\label{sec:iterative}

A salient feature of many reasoning tasks is an iterative component, meaning they can (or must) be solved by sequentially repeating certain operations. In this section, we use LEGO to study and compare Transformer architectures through the lens of iterative reasoning.

A natural solution to LEGO---and arguably the go-to solution for a human---is to implement a ``for-loop'', where each iteration resolves one step in the reasoning chain. The iteration could look for the next unresolved variable token whose value could be resolved in one step. 
%
Iterative Transformer architectures such as ALBERT and Universal Transformers~\cite{dehghani2018universal}, where the weights are shared across different layers, inherently implement a for-loop with a number of iterations equal to the number of layers. 
If the model manages to learn to implement one such iteration during training, the network would immediately be capable of performing length extrapolation. If this indeed occurs, it would point to a clear advantage of ALBERT over BERT in our setting. This leads to the following questions.


\subsubsection*{Q1. Do iterative architectures indeed exhibit better length extrapolation?}

The bottom plots of Figure \ref{fig:result_shortcut} display the length extrapolation result for BERT and for ALBERT. They show the clear advantage of recurrence: While the non-iterative BERT achieves only somewhat better-than-random accuracy for one variable (\#6) beyond the ones accounted for during training (\#0-
-\#5), the iterative ALBERT reaches near-perfect accuracy on two additional variables (\#6 and \#7), and nontrivial accuracy on the third (\#8). These results clearly support that iterative architectures do generalize better in the iterative LEGO reasoning task.

\subsubsection*{Q2.  Does the ALBERT architecture actually implement the for-loop?} 

To a lesser extent, Figure~\ref{fig:result_shortcut} also hints at a positive answer to Q2. Observe that ALBERT exhibits length extrapolation to variable \#6 immediately (in terms of epochs) as soon as it fits the training variables (\#0 -- \#5), whereas for BERT, the corresponding plot (\#6) climbs gradually even after the training variables are predicted perfectly. 
This suggests that once it manages to learn the operations required for one step of reasoning, it can immediately implement those operations over a few more iterations not required in training.

In order to gain stronger evidence, we measure the dependence between the location of a variable token in the chain and the layer in which its value is typically resolved. To this end, given a trained model, we train one linear classifier per layer which predicts the value of a variable token based only on its token representation at the corresponding layer (without using other information), while keeping the original model unchanged. 
This allows us to gauge the rate of information percolation along the reasoning chain in terms of layers per reasoning step. If the model indeed implements a for-loop in its forward pass, one expects a linear relationship between the number of layers and the number of reasoning steps already completed.
\begin{figure}[t] 
    \centering
    \includegraphics[trim={0cm 0cm 0cm 0cm},clip, width=15cm]{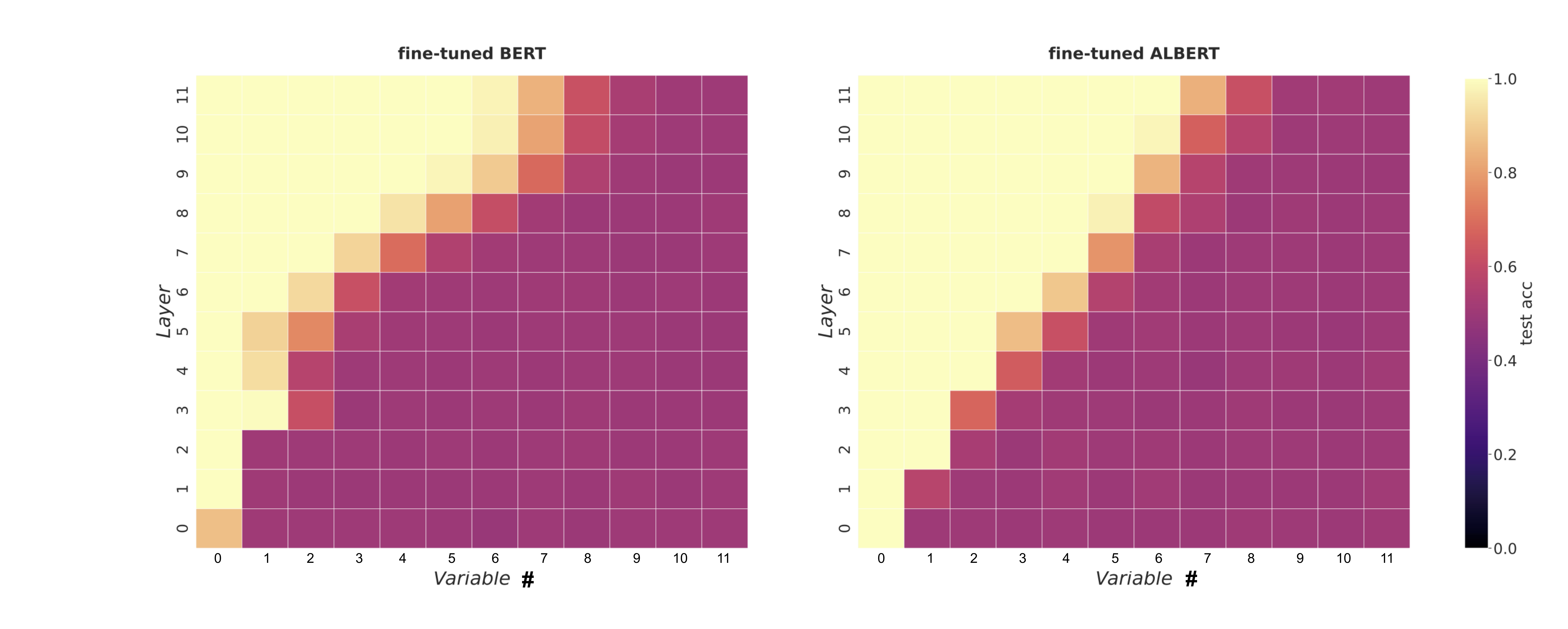}
    \caption{Visualization of information percolation within the fine-tuned models. The color indicates the test accuracy of the probing classifier at each layer. Brighter is higher. We observe ALBERT's information percolation is linear than BERT's, which implies ALBERT is biased towards learning a for-loop.}
    \label{fig:perco}
\end{figure} 
We visualize in Figure~\ref{fig:perco} the test accuracy of prediction as a function of the layer in the network and depth in the chain. 
While not perfectly linear, the relation clearly looks closer to linear in ALBERT, suggesting that the ALBERT model has an inductive bias towards learning to implement the ``natural" for-loop with its forward pass.  

\subsubsection*{Q3. How can we incentivize models to learn iterative solutions?} 

\begin{figure}[h]
     \centering
    \includegraphics[trim={0cm 0cm 0cm 0cm},clip, width=15cm]{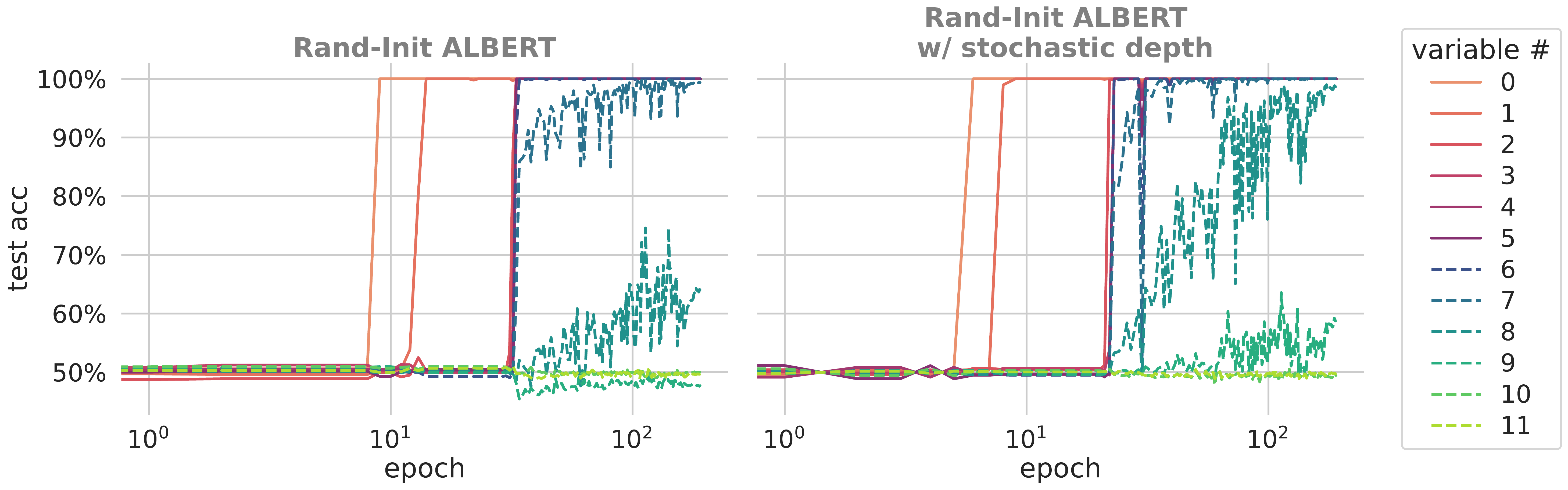}
    \caption{Generalization of ALBERT trained with stochastic depth. The stochastic depth improves the length extrapolation to longer sequence lengths.} 
     \label{fig:stochasticdepth}
\end{figure}

We attempt to incentivize the model to implement the ``natural" for-loop solution. We rely on the observation that if each iteration of the for-loop simply percolates the information one more step (assigning a value to the next variable), then adding more layers with the same weights should not affect the output, and in fact, one should be able to read out the output of the calculation from any layer of the neural network, as long as its depth exceeds the length of the chain. 
With this observation in mind, we train a ALBERT model with stochastic depth~\cite{huang2016deep}. We uniformly sample depth between 6 and 12 per batch during training while fixing it at 12 during test. Figure~\ref{fig:stochasticdepth} shows a clear improvement in generalization to longer lengths using stochastic depth.

\subsection{Rand-Init vs. Pretrained: Structural advantages from pretraining}
\label{sec:pretrain}
\begin{figure}[t]
    \centering
    \includegraphics[trim={0cm 0cm 0cm 0cm},clip, width=15cm]{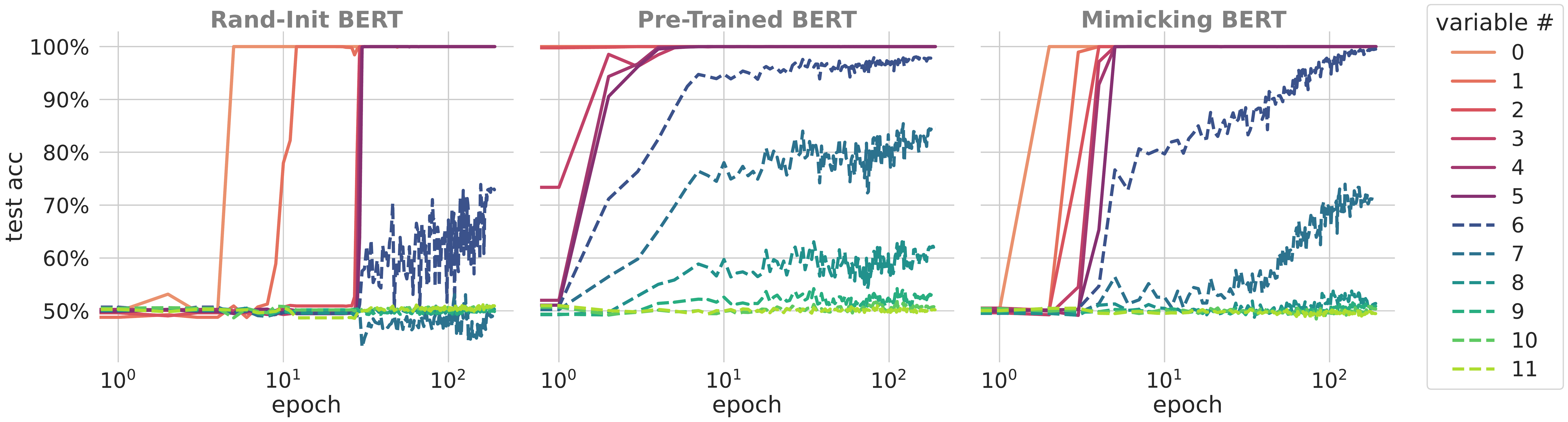}
    \caption{Pretrained BERT exhibits significant performance advantages over its Rand-Init counterpart, while the mimicking procedure (a simple initialization scheme we describe below) heads closes the gap.}
    \label{fig:pretrain}
\end{figure}

Pretraining large models has emerged as a prominent and highly successful paradigm in large-scale deep learning. It advocates first training the model on a large dataset to perform a generic task, followed by task-specific fine-tuning on the task at hand. Our goal here is to use LEGO as a testing ground for this paradigm. To this end, we compare (a) training the BERT architecture for LEGO from random initializations to (b) fine-tuning the standard pre-trained BERT model to solve LEGO. Figure \ref{fig:pretrain} (left and center plots) shows that pretraining helps generalization in LEGO dramatically: the pre-trained model generalizes to unseen sequence lengths (the dashed plots) much better, and within a far smaller number of epochs, than the randomly initialized model.

\subsubsection{Why does pretraining help in LEGO?}
One simple explanation is that pre-trained BERT is already aware of the semantics of tokens like `=' or `-'.
We have easily ruled out this possibility, by replacing those tokens with arbitrary ones that do not encompass the same semantics; this does not affect the performance of pre-trained BERT. A more intriguing explanation pertains to the attention mechanism itself. At its basis, LEGO requires two fundamental types of information transfer:
\begin{itemize}[topsep=0pt,itemsep=0ex,partopsep=0ex,parsep=0ex, leftmargin=2ex]
  \item \textbf{\emph{Association:}} encoding long-range dependencies that transfer a value between two occurrences of the same variable. For example, if the input contains the two clauses ``$a=+1$'' and ``$b=-a$'' (with arbitrary separation between them), the architecture must associate the two occurrences of the variable $a$ in order to correctly set $b$ to $-1$.
  \item \textbf{\emph{Manipulation:}} encoding short-range dependencies of transferring a value from the right-hand to the left-hand side of the clause. For example, to successfully process the clause ``$b=-a$'', the architecture must associate these particular occurrences of $a$ and $b$ with each other, in order to transfer the value of $a$ (after applying to it the group element $-1$) into $b$. 
\end{itemize}
Association corresponds to a purely global attention pattern, completely reliant on the \emph{identity} or \emph{content} of the tokens and oblivious to their \emph{positions} in the input sequence. Manipulation, in contrast, corresponds to a purely local attention pattern, where nearby positions attend to each other. 

\begin{figure}[h]
    \centering
    \includegraphics[trim={3cm 0cm 0cm 0cm},clip, width=15cm]{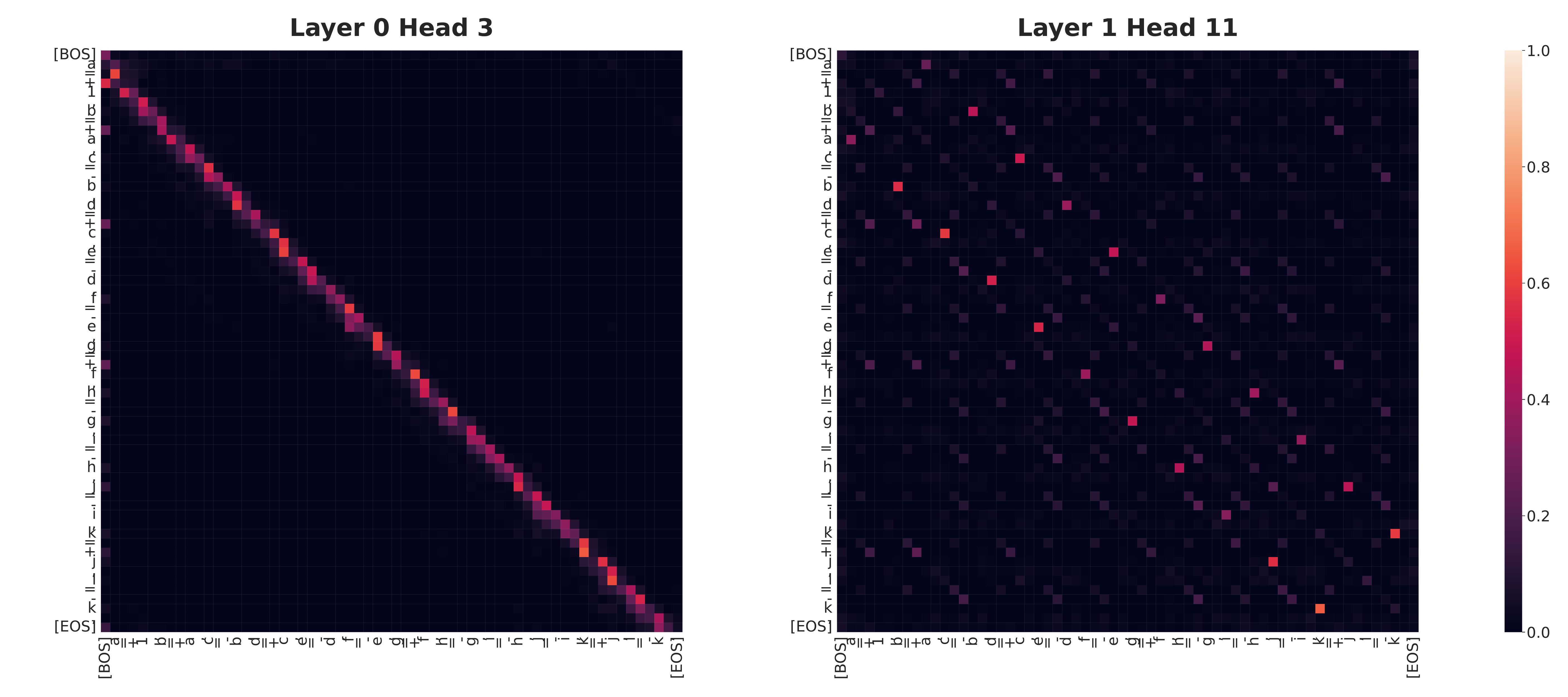}
    \caption{Visualization of two representative attention maps from a pre-trained BERT model not yet fine-tuned on LEGO. A complete visualization of all attention patterns of the pre-trained BERT is in Appendix~\ref{sec:attn-maps}. On the LEGO input sequence, certain heads implement local, convolution-like manipulation operators (left), while some others implement global, long-range association operators (right). Note that the sample input sequence is presented in the reasoning chain order for visualization purposes only.}
    \label{fig:attention_heads}
\end{figure}

It is natural to ask whether they are indeed manifested in the pre-trained model's attention heads in practice. Indeed, Fig.~\ref{fig:attention_heads} shows two exemplar attention heads of pre-trained BERT on an input LEGO sequence without any fine-tuning. The right head clearly depicts association: each token attends to all other occurrences of the same token in the input sequence.
This motivates us to make the following hypothesis: \emph{~\textbf{the advantage of pre-trained models on LEGO can be largely attributed to the association and manipulation heads learned during pretraining.}}

Note that merely the existence of the heads does not fully validate the hypothesis yet. To rule out other factors, we carefully design controlled experiments to test this hypothesis in the section below.

\subsubsection{Verifying the hypothesis with Mimicking}
 To test this hypothesis, we conduct the following \emph{mimicking} experiments.
 
\textbf{Mimicking BERT} ~We `initialize' certain attention heads to perform association and manipulation, without access to pretraining data. We achieve this by specifying the target attention matrices (one for association and one for manipulation), and training the model on random data to minimize a ``mimicking loss'' that measures how well the actual attention matrices at every layer match the target matrices. The precise mimicking loss and training protocol are specified in the Appendix~\ref{sec:mimic}. The rightmost plot in Figure \ref{fig:pretrain} shows that BERT with mimicking initialization attains significant advantage in generalization over randomly initialized BERT, despite not being pre-trained on any real data (and thus not having learned to ``reason''). This confirms that much of the advantage of pre-trained BERT stems from having learned these information transfer patterns.


\subsection{Shortcut solutions and their effect on generalization}
\label{sec:shortcut}
As discussed in Section~\ref{sec:iterative}, a natural solution to LEGO is to resolve variables iteratively by the order of their depth in the chain. Surprisingly, we find that the Rand-Init BERT and ALBERT models first learn a ``shortcut'' solution: they immediately resolve the \emph{last} variable in the reasoning chain, perhaps by counting the total number of minus signs. Indeed, the last variable can be easily identified as it appears only once whereas every other variable appears twice, 
and its value is fully determined by the parity of the number of minus signs. 
This behavior is observed in Figure~\ref{fig:shortcut}: the randomly initialized models are trained to fit all $12$ variables: the last variable (\#11, indicated by the brightest green curves) improves earlier than \emph{almost} all other ones.

This behavior may be related to the well-observed phenomenon of \emph{spurious features}: a model succeeds in training not relying on any actual features of cows and circumventing the intended solution~\cite{mccoy2019right, srivastava2020robustness, gururangan2018annotation, nguyen2021avoiding}. 

We use LEGO as a case study of shortcut solutions and their effect on generalization. Instead of training the model to fit the first six variables (as in bottom Figure~\ref{fig:result_shortcut} in Appendix), we train it to fit the first five (\#0--\#4) and the last variable (\#11). This allows us to measure length extrapolation (to \#5--\#10) in a setting 
where models can learn the shortcut.  
The results show significantly degraded performance, implying that shortcut solutions impede generalization. We then study ways to prevent models from learning them, by pretraining and mimicking. The full section appears in Appendix~\ref{sec:shortcut-full}.


\section{LEGO Attention: faster and better}
\label{sec:LEGO v0}
Our analysis in Section~\ref{sec:pretrain} reveals that the advantage of the pre-trained BERT model on LEGO originates from two specific types of attention structures emerging from pre-training --- the association and manipulation patterns. A quick examination of all the attention heads depicted in Appendix~\ref{sec:attn-maps} suggests that there is one more clearly identifiable attention pattern: broadcasting on the $\texttt{[CLS]}$ token or the $\texttt{[SEP]}$ token (sometimes both). Namely, it `broadcasts' the value inside the special tokens to the others. Even though $\texttt{[CLS]}$, $\texttt{[SEP]}$ play no role on LEGO per se, they are vital to the pretraining objective as well as many downstream tasks. Thus the broadcasting attention pattern is presumably important for many real-life NLP tasks beyond LEGO. Association, manipulation, and broadcasting consist of a considerable portion of the pre-trained BERT's attention heads, and they are so structured that we can in fact LEGO v0 them efficiently. 

\begin{figure}[H]
\centering
\includegraphics[trim={0cm 0cm 0cm 0cm},clip, width=8cm]{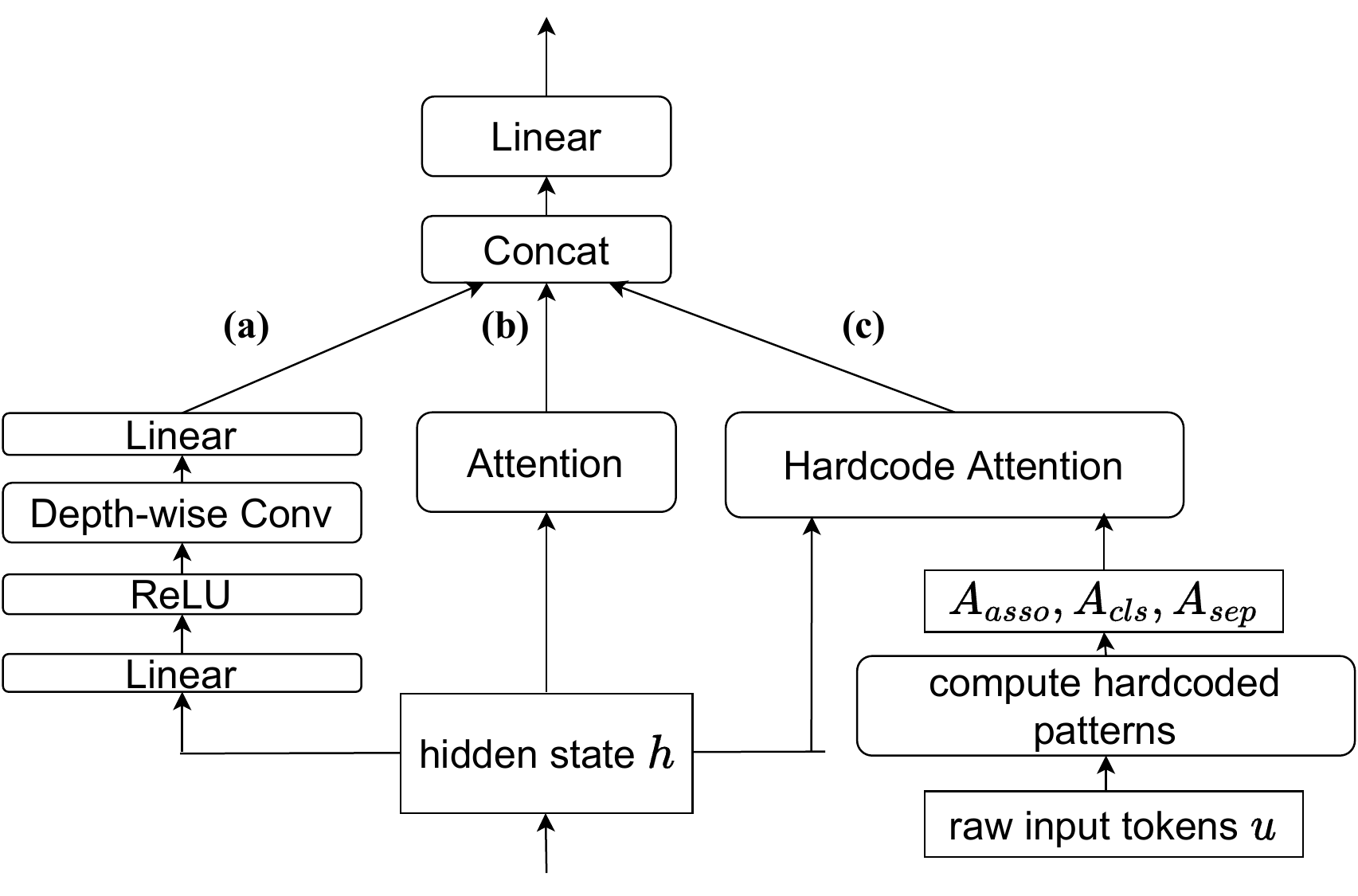}
\caption{Our proposed LEGO attention consists of $3$ pathways. BERT has pathway (b) only; the LEGO v0 attention module has (a) and (c); the LEGO v1 attention has (a), (b), and (c). See Appendix~\ref{sec:app-LEGO v0}. }
\end{figure}

\paragraph{LEGO Attention:}~~For the association, manipulation, and broadcasting heads, we can efficiently construct the sparse attention matrix based on the input token IDs only, without learning $Q$ and $K$ or the expensive attention probability computation. For manipulation maps, due to their intrinsic locality, we decide to implement them directly with temporal convolutions (along the time dimension). For the other global maps, given a \emph{raw} input sequence of $T$ tokens, $u_1, u_2, \ldots, u_T\in\mathbb{N}$, we manually construct the association and broadcasting maps $A_{asso}$, $A_{cls}$, $A_{sep}\in\mathbb{R}^{T\times T}$ such that  $(A_{asso})_{ij} = \mathbf{1}\left[u_i=u_j\right], ~(A_{cls})_{ij} = \mathbf{1}\left[u_j=\texttt{[CLS]}\right], ~(A_{sep})_{ij} = \mathbf{1}\left[u_j=\texttt{[SEP]}\right]$
where $\mathbf{1}\left[\cdot\right]$ is the indicator function which outputs $1$ if the argument is true and $0$ otherwise. In the end, we normalize them to have row-wise unit $\ell_1$ norm. Notably, the latter three patterns require no training (except for a value map for each layer) and are shared across all layers.

\begin{figure}[t]
    \centering
    \includegraphics[trim={0cm 0cm 0cm 0cm},clip, width=16cm, height=4cm]
    {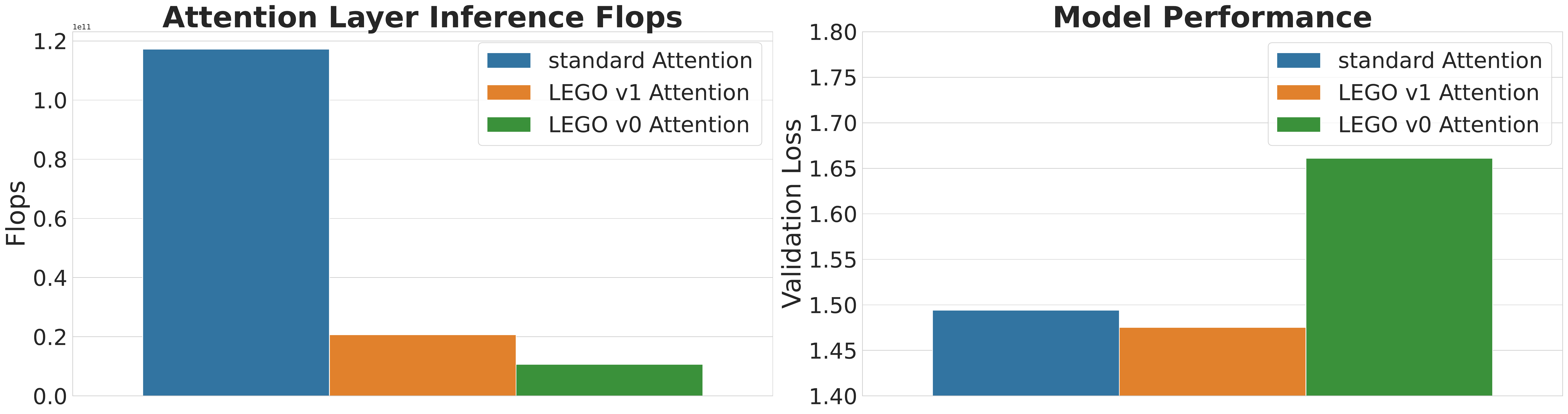}
    \caption{\textbf{Top)} Comparison of inference Flops and model size. Flops are measured on a batch of $64$ sequences of 512 tokens.}

    \includegraphics[trim={0cm 0cm 0cm 0cm},clip, width=16cm]{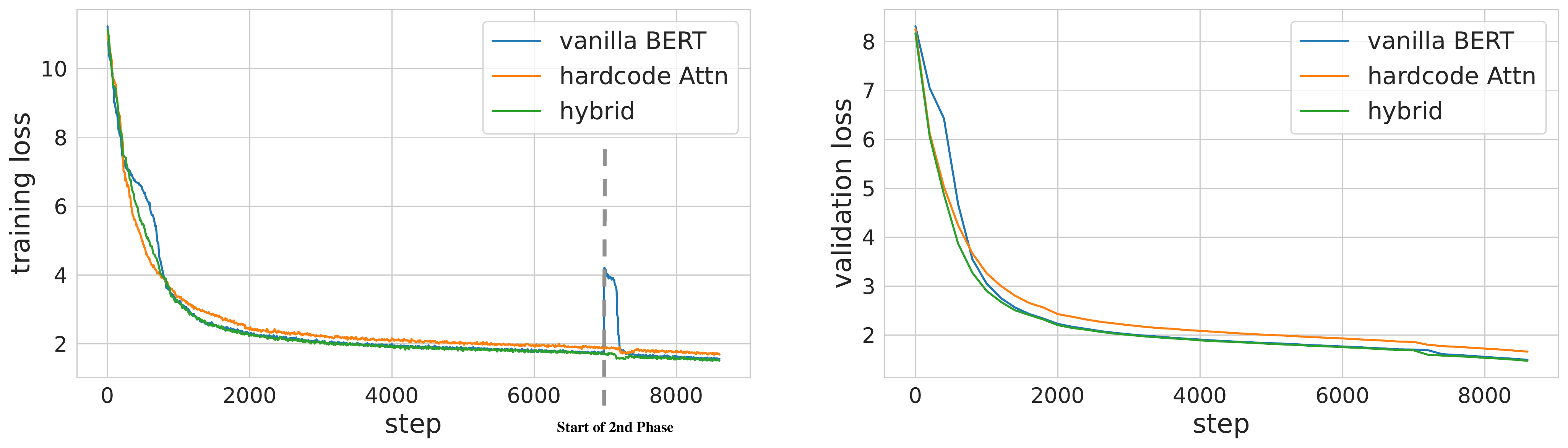}

    \caption{ Training and validation performance on BERT pertaining task (Masked Language Modelling+Next Sentence Prediction). As a standard, the training sequence length increases from 128 to 512 around the 7k-th step, where the BERT training loss exhibits a sudden bump in response, while the LEGO v0/v1 models exhibit remarkable resilience. The LEGO v1 model learns faster and (slightly) outperforms BERT in validation. }
    \label{fig:hybrid_pretrain}
\end{figure}

On the standard BERT pertaining benchmark, we compare the following three models: BERT-base model, LEGO v0 and v1 models. We use convolutional kernel size 21 for the latter two.
In Figure~\ref{fig:hybrid_pretrain}, we show that the LEGO v0 model learns fast in the beginning but falls short later on. However, the LEGO v1 model not only reduces model size and accelerates inference, but also renders models that are extremely competitive with the base model in terms of the final performance of large-scale pertaining. We follow precisely the training pipeline and hyperparameters of~\cite{devlin2018bert}. See Appendix~\ref{sec:app-LEGO v0} for architecture details of the LEGO v0/v1 models.

We observe that the LEGO v0 model learns faster but gradually falls short, while the LEGO v1 model achieves the best of both worlds: it learns faster at the beginning and achieves even (slightly) lower validation loss at the end. The LEGO v1 model's validation loss curve appears to be a lower envelope of the other two. The BERT/LEGO v0/v1 models achieve $1.49/1.66/1.47$ final pertaining validation loss and $88.2/82.5/88.1$ Dev F1 score on SQuAD v1.1~\cite{rajpurkar-etal-2016-squad}. We leave comprehensive evaluations for future work. 

\section{Conclusion}

In this work, we study Transformers by constructing LEGO, a controllable synthetic logical reasoning task. With LEGO, we have gained insights into their inductive bias, the role of pertaining, etc. Based on these insights, we proposed the LEGO attention mechanism which both accelerates inference and leads to comparable or even better performance. There are many important attention heads beyond just manipulation and association, and their roles remain to be discovered. We believe LEGO will continue to deepen our understanding on Transformers' inner working and to inspire better algorithms/architectures for tasks beyond LEGO. 




\clearpage



\bibliographystyle{alpha}
\bibliography{dlbib}

\newpage
\appendix
\section{Shortcut solutions and their effect on generalization}\label{app:shortcut}
\label{sec:shortcut-full}
As explained in Section~\ref{sec:shortcut}, we have observed that the randomly initialized models first learn a ``shortcut'' solution---predicting the last variable in the chain by counting the overall number of minus signs---instead of the ``common sense'' iterative solution. This can be seen in the top two plots in Figure~\ref{fig:result_shortcut}, where the accuracy of variable \#11 improves earlier than most of the other variables.\footnote{This behavior is not expected in the bottom two plots, since in the top ones the models are trained to fit all 12 variables including \#11, while in the bottom plots the models are only trained to fit the first 6 variables, which precludes learning the shortcut solution.}

To be more precise, let us describe the two solutions in detail with an example. Consider the input:  $\texttt{a=+1;\,d=-c;\,b=-a;\,c=b;}$
Initially, only the variable $\texttt{a}$ is resolved. The iterative solution identifies an unresolved variable that appears in the same clause with an already resolved variable, resolves it according to that clause, and repeats. In this example, it would resolve \texttt{b} to $-1$ by the clause \texttt{b=-a}, then resolve \texttt{c} to $-1$ by the clause \texttt{c=b}, and then resolve \texttt{d} to $1$ by the clause \texttt{d=-c}. The shortcut solution identifies an unresolved variable that appears only once, and resolves it to $1$ if the overall number of minus signs is even, and to $-1$ otherwise. In the above example, where \texttt{d} is the last variable in the reasoning chain, the shortcut solution correctly resolves it to $1$.

As further mentioned in Section~\ref{sec:shortcut}, the shortcut solution to LEGO may be related to the phenomenon of spurious features, where models learn to perform tasks in ways that circumvents the intended ``common sense'' solution a human would use. Such spurious solutions are often considered undesirable, as they are known to generalize poorly even to mild variants of the task.
Indeed, the shortcut solution to LEGO is brittle even under simple variations to the problem:
\begin{CompactItemize}
    \item Repeated clauses, e.g., \texttt{a=+1;\,b=-a;\,b=-a;}
    \item Redundant clauses, e.g., \texttt{a=+1;\,b=a;\,c=-a;\,c=-b;}
    \item Multiple jointly rooted reasoning chains, e.g., \texttt{a=+1;\,b=-a;\,c=-a;}
    \item Multiple disjoint reasoning chains, e.g.,  \texttt{a=+1;\,b=+1;\,c=-a;\,d=-b;}
\end{CompactItemize}
and more. In all those settings the shortcut solutions would fail, whereas the ``common sense'' iterative solution would succeed.
This motivates us to empirically study the effect of the shortcut solution on the ability of the models to generalize. 
We pose the following questions:

\begin{figure}[H]
    \centering
    \begin{subfigure}{\textwidth}
     \begin{minipage}[r]{0.05\textwidth}
     \caption{}
  \end{minipage}
  \begin{minipage}[l]{0.95\textwidth}
    \includegraphics[trim={0cm 0cm 0cm 0cm},clip, width=12cm]{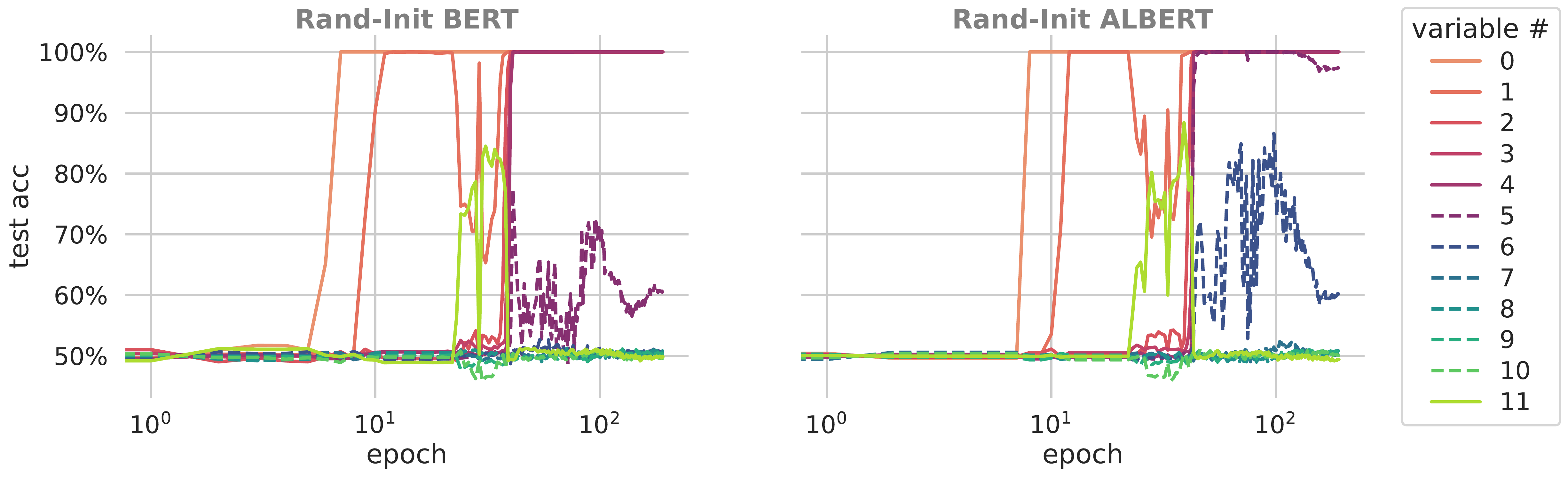}
    \end{minipage}
    \end{subfigure}
    \vspace{0.1cm}
    \hrule
    \vspace{0.1cm}
\begin{subfigure}{\textwidth}
     \begin{minipage}[r]{0.05\textwidth}
     \caption{}
  \end{minipage}
  \begin{minipage}[l]{0.95\textwidth}
    \includegraphics[trim={0cm 0cm 0cm 0cm},clip, width=12cm]{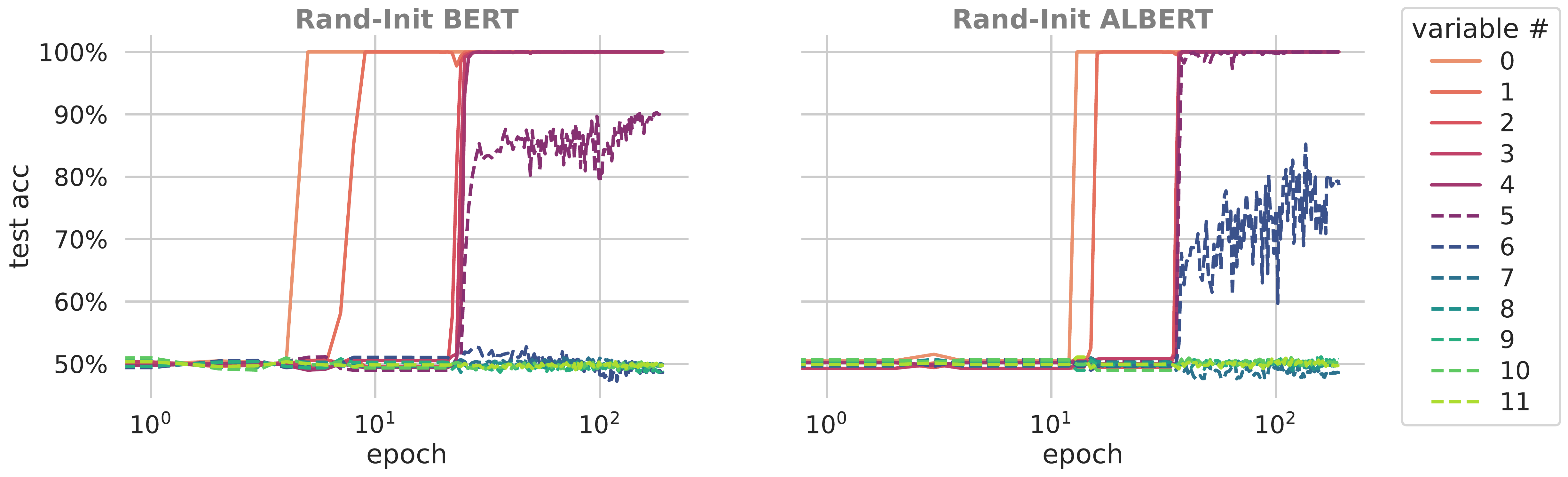}
    \end{minipage}
    \end{subfigure}
    \caption{Learning shortcut impedes generalization. (a) Train on variables \#0-\#4 and \#11. (b) Train on variables \#0-\#4 only. In both plots we test on all 12 variables. }
    \label{fig:shortcut-gen}
\end{figure}

\subsubsection*{Q1. How does reliance on shortcut solutions affect the ability of the network to generalize?}
A first indication that the shortcut solution is undesirable for LEGO can be gleaned already from Figure~\ref{fig:result_shortcut}: along with the early improvement in the accuracy of variable \#11 (which indicates that the shortcut solution is being learned), we observe a drop in the accuracy of some of the variables that were already learned (\#2 in ALBERT and \#3 in BERT). This may suggest that the shortcut solution impedes even classical generalization. Indeed, the models seem to ``realize'' that, as we can see that the accuracy of \#11 drops before improving again together with the other variables, indicating that the shortcut solution is being abandoned in favor of a solution that can predict all variables correctly.

To gain insight into the effect of the shortcut solution on out-of-distribution generalization, we performed an experiment where the models are trained to fit the first five variables (\#0-\#4) and the last one (\#11), and are asked to predict all 12 variables at test time. This is different from the top plots in Figure~\ref{fig:result_shortcut} where the model was trained to fit all 12 variables (and thus no out-of-distribution generalization is observed), and from the bottom plots in Figure~\ref{fig:result_shortcut} where the model is trained to fit the first six variables (\#0-\#5), without \#11 (and thus learning the shortcut solution is not possible). The results of this experiment are reported in the top two plots in Figure~\ref{fig:shortcut-gen}. The bottom plots depict a control experiment where the models are trained to fit only the first five variables, without \#11 (and thus, again, learning the shortcut solution is not possible). The results show that the models exhibit inferior out-of-distribution generalization (to variables \#5 in BERT and \#6 in ALBERT) when provided supervision for \#11 (top plots), even though they are given strictly more information during training than in the bottom plots, ostensibly making the task easier. We thus infer that the shortcut solution in LEGO has an adverse effect on out-of-distribution generalization.

\begin{figure}[tb] 
    \centering
    \includegraphics[trim={0cm 0cm 0cm 0cm},clip, width=13cm]{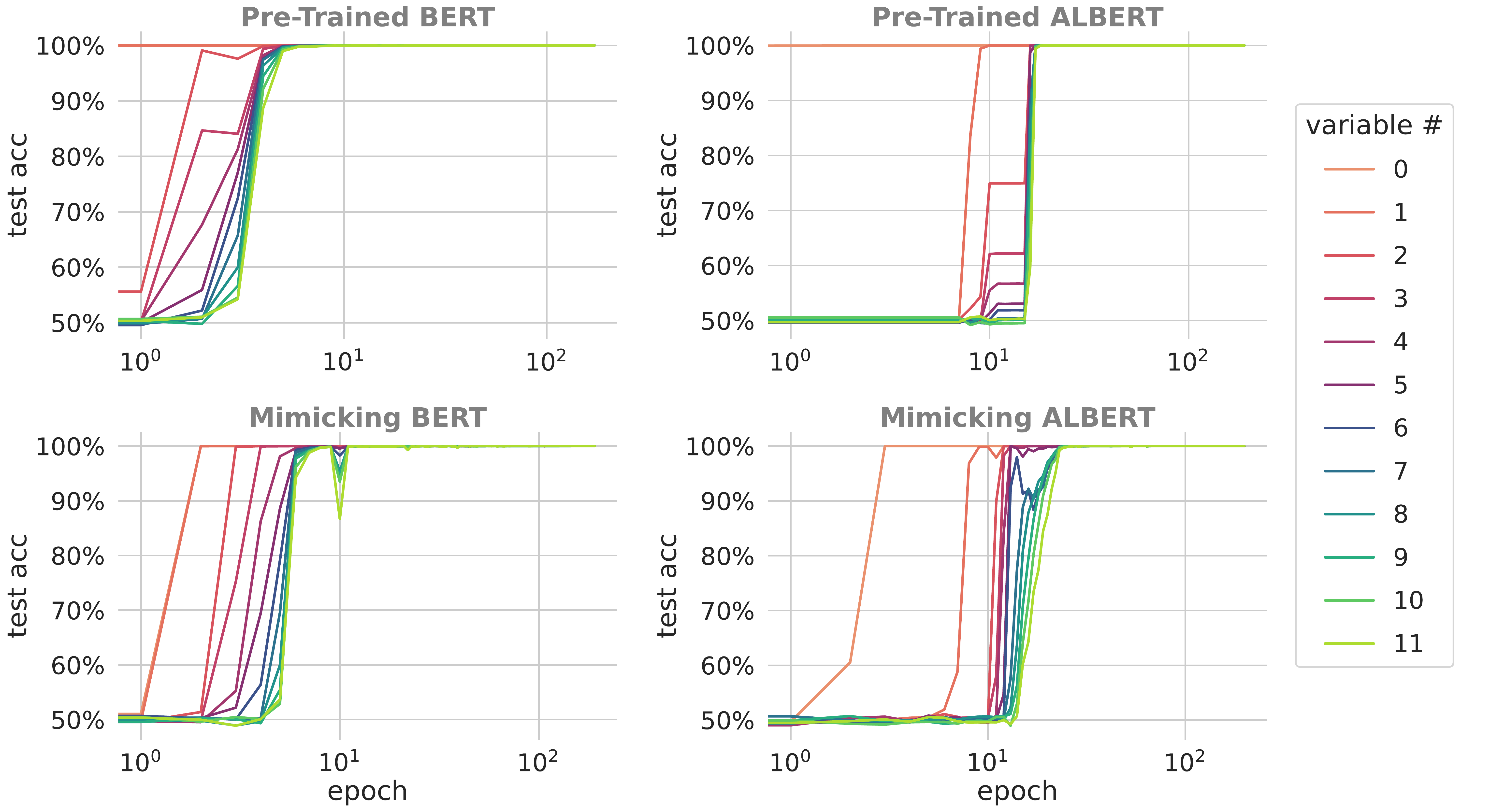}
    \caption{Shortcut solutions are avoided via either pre-training or mimicking. }
    \label{fig:avoid-shortcut}
\end{figure}

\subsubsection*{Q2. What are effective ways to prevent models from learning shortcuts, and do they result in better generalization?}

In Section~\ref{sec:4} 
we studied the effect of pretraining on LEGO, and observed that the pretrained BERT model exhibits much better out-of-distribution generalization than the randomly initialized BERT model. This naturally suggests that pretrained BERT possibly avoids the shortcut solution. 
We confirm experimentally in Figure~\ref{fig:avoid-shortcut} (top), where pretrained BERT and ALBERT are finetuned to fit all 12 variables. Indeed observe the accuracy of \#11 improving either later than or concurrently with all other variables, suggested that the shortcut solution is not being learned. We speculate that this may have to do with the number of epochs it takes to learn the iterative solution: By the time the randomly initialized BERT has learned the shortcut solution, pretrained BERT already attains full accuracy on the entire chain of variables. Avoiding the shortcut solution may partly explain the superior out-of-distribution generalization performance of pretained BERT over randomly initialized BERT, seen in Figure~\ref{fig:pretrain}.

Section~\ref{sec:4} also showed that our mimicking technique---which directly mimics the attention patterns of the pretrained model without training on any data---can recover much of the benefit in pretraining for LEGO. This extends to avoiding the shortcut solution as well: the bottom plots in Figure~\ref{fig:avoid-shortcut} show that the mimicking BERT and ALBERT models exhibit similar accuracy patterns to their pretrained counterparts, suggesting that the shortcut solution is not being learned by them. 


\section{Data generation and training details for the LEGO task} 
\subsection{Data generation}  
\label{sec:data-generation}
We  specify the data generation mechanism for Task 1 in the following. We use lowercase alphabets as variables $A=\{a,b,c,d,\ldots, z\}$. Given $n$, we generate a sentence from our distribution $s\sim \mathcal{D}(n)$ as follows: 
\begin{CompactItemize}
\item[1.] Sample $n$ variables $a_1,a_2,\ldots,a_n \in A$ and their corresponding assignments (or labels) $y_1,y_2,\ldots y_n\in X$ uniformly, i.e., $\forall\,i\in[n]$, $a_i\sim\text{Unif}(A)$ and $y_i=\pm 1 \text{ w.p. } 0.5$.
\item[2.]  The $n$ clauses are then generated as $a_i=g_i a_{i-1}$ for $i=1,2,\ldots, n$, where $a_0=r=1$ and the group elements $g_1,g_2,\ldots g_n\in G$ are uniquely chosen so that the clauses are consistent with assignments $a_i=y_i$ for all $i\in[n]$.
\item[3.] The sentence $s$ is generated by concatenating a random ordering of the $n$ clauses with a semicolon $;$. Finally, the sentence is padded with \texttt{[BOS]} and \texttt{[EOS]} tags to denote the beginning and end of sentence, respectively. See Figure~\ref{fig:data_sample_task1} for example sentences from our distribution. 
\end{CompactItemize}

\begin{figure}[htb]
\centering
\begin{scriptsize}
\begin{verbatim}
        [BOS] j=-f; f=-b; y=+t; o=+e; d=+y; v=+d; h=-o; b=-i; i=+1; t=+l; e=-j; l=-h; [EOS]
        [BOS] j=+o; s=-y; p=-r; y=-m; u=-a; a=-f; k=+p; o=-k; q=+u; m=+1; f=+s; r=+q; [EOS]
        [BOS] z=+d; b=+1; m=+t; d=-u; u=-h; a=-b; j=+m; i=-j; t=+x; f=+i; h=-f; x=-a; [EOS]
        [BOS] j=-f; f=-b; y=+t; o=+e; d=+y; v=+d; h=-o; b=-i; i=+1; t=+l; e=-j; l=-h; [EOS]
        [BOS] w=+l; m=+c; c=-i; f=-d; p=-m; a=+b; y=-a; b=+p; i=+f; l=-v; d=+1; v=+y; [EOS]
\end{verbatim}
\end{scriptsize}

\caption{Samples of sentences generated from our distribution $\mathcal{D}(n)$ for Task $1$ with $n=12$. 
}
\label{fig:data_sample_task1}
\end{figure}

\subsection{Training} \label{training}Our vocabulary for data generated as above thus consists of symbols of variables $a\in A$, group operations $+,-\in G$, the root node $1\in X$, the equal sign `=', and the semicolon `;' along with the \texttt{[BOS]} and \texttt{[EOS]} tags. To apply transformers to the LEGO task we convert each symbol in our vocabulary to vectors in $\R^d$ (referred to as {\em tokens}) using a learnable linear embedding layer. Thus, a sentence of $n$ clauses now corresponds to an ordered list of $5n+2$ tokens, which we turn into an unordered list using positional embedding (see \cite{Vas17}). These tokens are processed iteratively by {\em transformer blocks}. Each transformer block maps $5n$ tokens in $\mathbb{R}^d$ to another set of $5n$ tokens using a multi-head attention layer followed by a one-hidden layer feedforward net (there are also residual connections and layer normalization, for full details of architecture see \cite{Vas17}). 
We use \texttt{bert-base-uncased}\footnote{\url{https://huggingface.co/bert-base-uncased}} and \texttt{albert-base-v1}\footnote{\url{https://huggingface.co/albert-base-v1}} along with their pretrained weights from the open source Huggingface transformers library~\cite{wolf-etal-2020-transformers}. The Rand-Init models have identical configurations to their pretrained counterparts but randomly initialized weights.

Our training and test datasets are i.i.d. samples from our distribution $D(n)$ as described above. We generate $10^4\times n$ and  $10^3\times n$ datapoints for training and test, respectively, and sanity checked that there is no overlap between train and test data. Recall that during training we provide supervision on the first $n_{tr}$ appearing in the graph representation of the sentence, but test the accuracy on all $n$ samples at test time. Note that since the clause positions are randomized in our input sentence (e.g., Figure~\ref{fig:data_sample_task1}), the first $n_{tr}$ clauses in the graph representation can appear at arbitrary positions in the sentence which allows for training the positional encodings for longer sequences than those seen in training.

In all the LEGO experiments, we use cross entropy loss averaged over the $n_{tr}$ clauses as our sample loss during training. We train for $200$ epochs using the Adam optimizer with batch size of $1000$ samples, $5\times10^{-5}$ learning rate, $\beta_1=0.9$, $\beta_2 = 0.999$, $\epsilon=1\times10^{-8}$, and cosine learning rate schedule with $T_{\max}=200$. For tokenization, we use  the pre-trained BERT tokenizer for all the experiments, which merely converts the input symbols into integers that are used as token ids. Each run is conducted on a cluster with $4$ A100 GPUs.

\begin{figure}[htb]
    \raggedright
    \begin{minipage}[l]{0.9\textwidth}
    \begin{subfigure}{\textwidth}
        \begin{minipage}[r]{0.05\textwidth}
        \caption{}
        \end{minipage}
        \begin{minipage}[l]{0.95\textwidth}
            \centering
            \includegraphics[trim={0cm 0cm 0cm 0cm},clip, width=12cm]{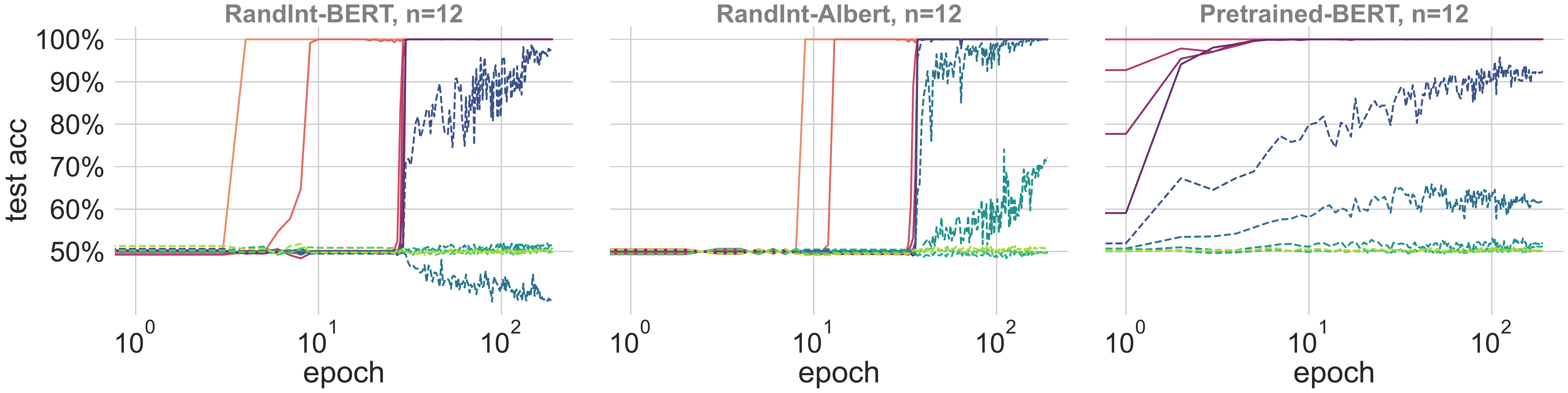}
        \end{minipage}
    \end{subfigure}
    \vspace{0.1cm}
    \begin{subfigure}{\textwidth}
        \begin{minipage}[r]{0.05\textwidth}
        \caption{}
        \end{minipage}
        \begin{minipage}[l]{0.95\textwidth}
            \centering
            \includegraphics[trim={0cm 0cm 0cm 0cm},clip, width=12cm]{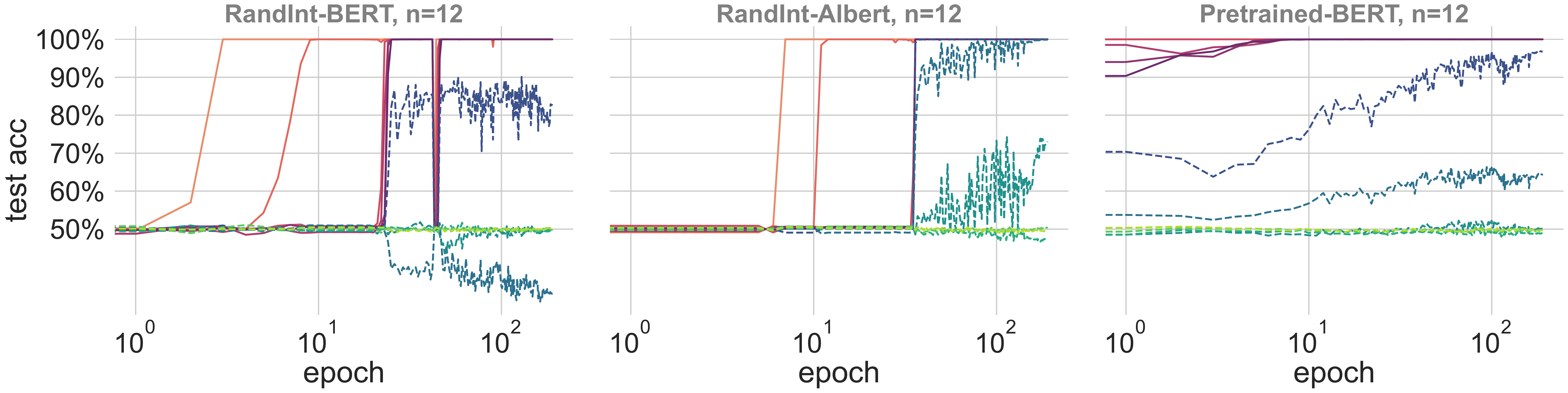}
        \end{minipage}
    \end{subfigure}
    \begin{subfigure}{\textwidth}
        \begin{minipage}[r]{0.05\textwidth}
        \caption{}
        \end{minipage}
        \begin{minipage}[l]{0.95\textwidth}
            \centering
            \includegraphics[trim={0cm 0cm 0cm 0cm},clip, width=12cm]{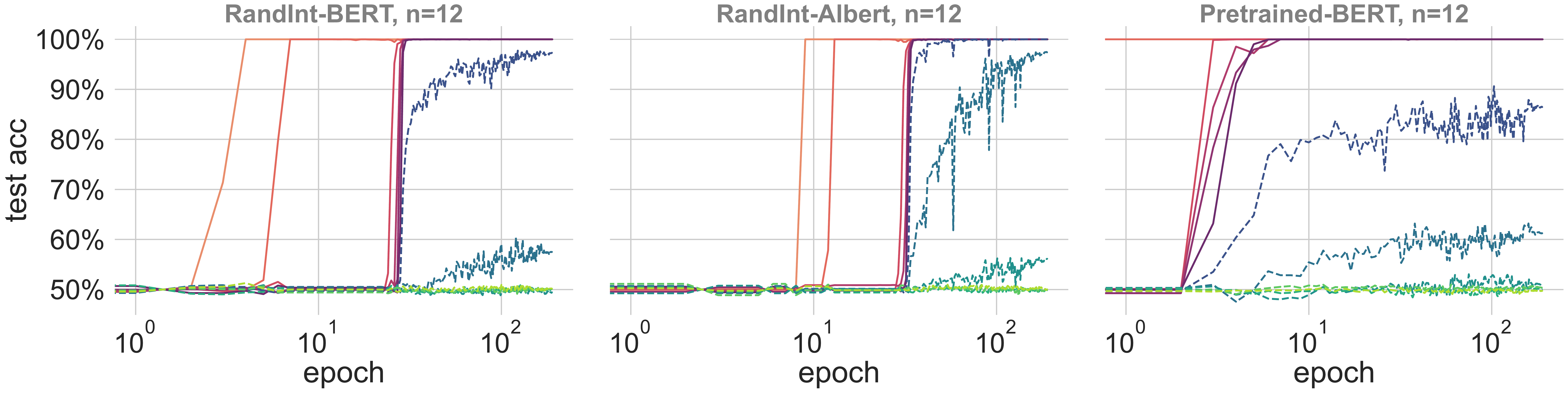}
        \end{minipage}
    \end{subfigure}
    \end{minipage}
    \begin{minipage}[l]{0.05\textwidth}
        \centering
        \includegraphics[trim={0cm 0cm 0cm 0cm},clip, width=1cm]{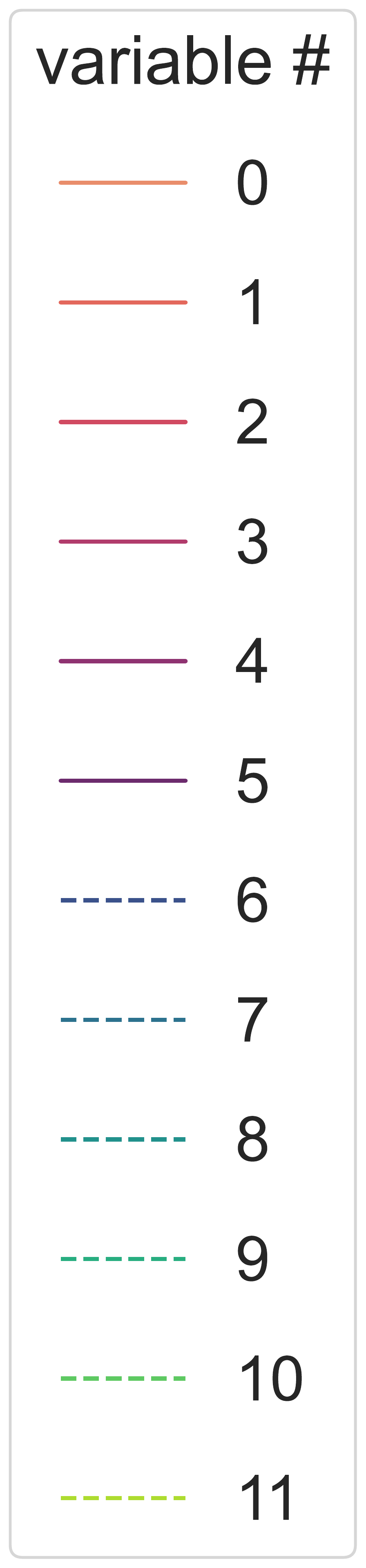}
    \end{minipage}
    \caption{Three sample runs of models trained on LEGO tasks with $n=12$ and $n_{tr}=6$. We observe that while there are variances across different runs of the models,  the qualitative conclusions stated in the paper holds for all the runs: i.e, iterative ALBERT models and pretraining lead to better generalization to unseen task lengths. }
    \label{fig:variance}
\end{figure}

\paragraph{A note on variance of training across LEGO tasks} 
When training BERT and ALBERT models for the LEGO task using our experimental setup, we see non-trivial variance in absolute test accuracies across different runs. In Figure~\ref{fig:variance}, we show three different runs of BERT and ALBERT models trained on LEGO tasks of length $n=12$ (the configuration used in most results in the paper). While we see that the absolute values of the test accuracies vary significantly, the qualitative observations made in our paper hold across all the runs: importantly, across all the runs, we see  that using iterative ALBERT architecture as well as pre-training non-iterative BERT architecture leads to better generalization to unseen lengths. This shows that conclusions derived in our paper hold despite the variance across runs.

Methodologically, we attribute the variance in the absolute test accuracy to the relatively small size of our datasets (e.g., the number of training examples for $n=12$ is $120K$ tokens) for training standard trained language models which are otherwise trained on hundreds of millions of tokens. Furthermore, note that in our experiments the variance across runs arise both from  having different train and test datasets as well as different random seeds for model initialization and training algorithm.

\subsection{The Mimicking procedure} 
\label{sec:mimic}
Starting with a randomly initialized transformer model, we would like to make sure that some of its attention heads implement the manipulation and association functionalities, \emph{prior to} the fine-tuning. To do so, we craft the desired attention patterns of a manipulation head and an association head, and train the model with gradient descent to match them. Specifically, given a random input $x\in\R^{T}$ of length $T$, we hard code the following matrix $M\in\R^{T\times T}$ as the target attention pattern for the manipulation head\footnote{Here we choose the Gaussian filter $[1,2,4,2,1]$ for illustration. In practice, we find that the final performance is robust to various pattern choices as long as they are localized and shift-invariant.}, and derive from input $x$ the following matrix $A\in\R^{T\times T}$ whose $A_{i,j}$ entry indicates whether $x_i$ and $x_j$ are identical tokens. Note that we further specify that $A$ has a zero diagonal in observance of the association head in Figure~\ref{fig:attention_heads} having a vanishing diagonal. In reality, attention maps have unit row sums, thus we normalize $M$ and $A$ accordingly to obtain $\Tilde{M}$ and $\Tilde{A}$ such that their rows $\Tilde{M}[t,:]$ and $\Tilde{A}[t,:]$ are valid distributions.

\begin{align*}
M = \begin{bmatrix}
&\ddots &\ddots&  \ddots& &   &   & & &  \\
& 1 & 2 & 4 & 2 & 1 &   &  & &  \\
& & 1 & 2 & 4 & 2 & 1 &   &  & \\
&  & & 1 & 2 & 4 & 2 & 1 &  & \\
& & &    & & \ddots     & \ddots &  \ddots & &
\end{bmatrix},~~~
    A_{ij}=\begin{cases}1~,~~~\text{if}~ x_i=x_j~\text{and}~i\neq j\\
    \\
0~,~~~\text{otherwise}\end{cases}
\end{align*}

At every layer, we randomly appoint two attention heads to mimic the manipulation and association operators while leave the other heads off the hook. Upon seeing a input sequence $\mathbf{x}\in\R^T$, we denote the attention maps of the appointed heads at the $l-$the layer as $\mathtt{Attn}_0^{(l)}(\mathbf{x}), \mathtt{Attn}_1^{(l)}(\mathbf{x})\in\R^{T\times T}$. For the mimicking objective, we draw input sequence $\mathbf{x}$'s whose tokens are independent and uniform over the vocabulary, and then compute the the Kullback–Leibler divergence between each row of $\mathtt{Attn}_0^{(l)}(\mathbf{x}), \mathtt{Attn}_1^{(l)}(\mathbf{x})$ and the corresponding rows of $\Tilde{M}$, $\Tilde{A}$. Thus the overall mimicking loss is

\begin{align*}
  L_{\mathtt{mimic}}  = \underset{\text{rand. seq.}~ \mathbf{x}}{\E}\left[\sum_{l=0}^{L-1} \frac{1}{T}\sum_{t=1}^{T} \left[\mathbf{KL}\left(\mathtt{Attn}_0^{(l)}(\mathbf{x})[t, :] \bigg\|\Tilde{M}[t, :]\right) + \mathbf{KL}\left(\mathtt{Attn}_1^{(l)}(\mathbf{x})[t, :] \bigg\|\Tilde{A}[t, :]\right) \right]\right]
\end{align*}

Note that the above mimicking loss pertains only two attention heads per layer and we leave out all the rest. Then the mimicking procedure boils down to updating the transformer model's parameters to minimize the  $L_{\mathtt{mimic}}$. We find that a vanilla Adam optimizer drives the mimicking loss down to near zero in a negligible amount of time compared to large-scale pre-training, even for large models such as BERT and ALBERT.

\section{Details of the LEGO v0/v1 models}
\label{sec:app-LEGO v0} 
The LEGO v0/v1 models are derived from the BERT-base model. They have same number of layers, same hidden dimensions, and same feed-forward layer's structure as the BERT model while and they differ in attention layers. In all three models, each attention head effectively operates on 64 hidden dimensions. The specifics are provided below. 
\begin{CompactItemize}
\item BERT-base model has 12 layers, 768 hidden dimensions, and 12 attention heads per layer.
\item LEGO v0 model: convolutional pathway consists of Linear from 768 dim to 576 dim + ReLU + Depthwise temporal Convolution with with kernel size 21 and 576 channels + Linear from 576 dim to 576 dim. There are three hardcoded attention heads whose attention patterns are $A_{asso}, A_{cls}, A_{sep}$ and each of them operates on a value map of 64 dim (thus the entire value map is a Linear layer from 768 dim to 192 dim).
\item LEGO v1 model: convolutional pathway consists of Linear from 768 dim to 384 dim + ReLU + Depthwise temporal Convolution with kernel size 21 and 384 channels + Linear from 384 dim to 384 dim. There are three hardcoded attention heads whose attention patterns are $A_{asso}, A_{cls}, A_{sep}$ and each of them operates on a value map of 64 dim (thus the entire value map is a Linear layer from 768 dim to 384 dim). There are also 3 ordinary attention heads whose Q,K,V maps are all Linear layers from 768 dim to 384 dim.
\end{CompactItemize}

\section{Effect of length of LEGO chains and depth of model}

All our experiments in the main paper were on a typical instance of LEGO task with length $n=12$ chains and standard BERT and ALBERT models of depth $D=12$. 
In this Appendix, we briefly explore the effect of the LEGO chain length $(n,n_{tr})$ and transformer models depth $D$. 
The chain structure of information flow in our LEGO Task 1 would suggest that for a transformer network of depth $D$, the learning and generalization on LEGO task would crucially depend on its maximum chain length $n$. If $n$ (or importantly $n_{tr}$) is too small, the training data might not have enough information to guide generalization to longer lengths, on the other hand, if $n$ is too large, the model might not be able to propagate information along the chain in a natural iterative manner. For example, implementing a ``natural" iterative algorithm of  resolving one clause of the task at a time (as described in the beginning of Appendix~\ref{app:shortcut}) would require models of depth $D\ge n$. 

\begin{figure}[H]
    \centering
    \begin{subfigure}{\textwidth}
        \begin{minipage}[r]{0.1\textwidth}
        \caption{$n=8$}
        \end{minipage}
        \begin{minipage}[l]{0.85\textwidth}
            \centering
            \includegraphics[trim={0cm 0cm 0cm 0cm},clip, width=13cm]{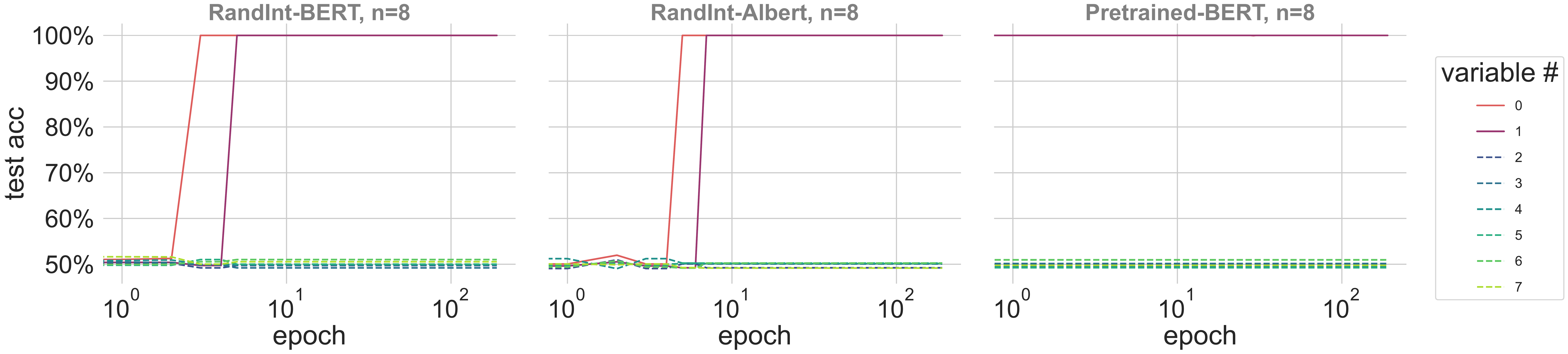}
        \end{minipage}
    \end{subfigure}
    \vspace{0.1cm}
    \hrule
    \vspace{0.1cm}
    \begin{subfigure}{\textwidth}
        \begin{minipage}[r]{0.10\textwidth}
        \caption{$n=12$}
        \end{minipage}
        \begin{minipage}[l]{0.85\textwidth}
            \centering
            \includegraphics[trim={0cm 0cm 0cm 0cm},clip, width=13cm]{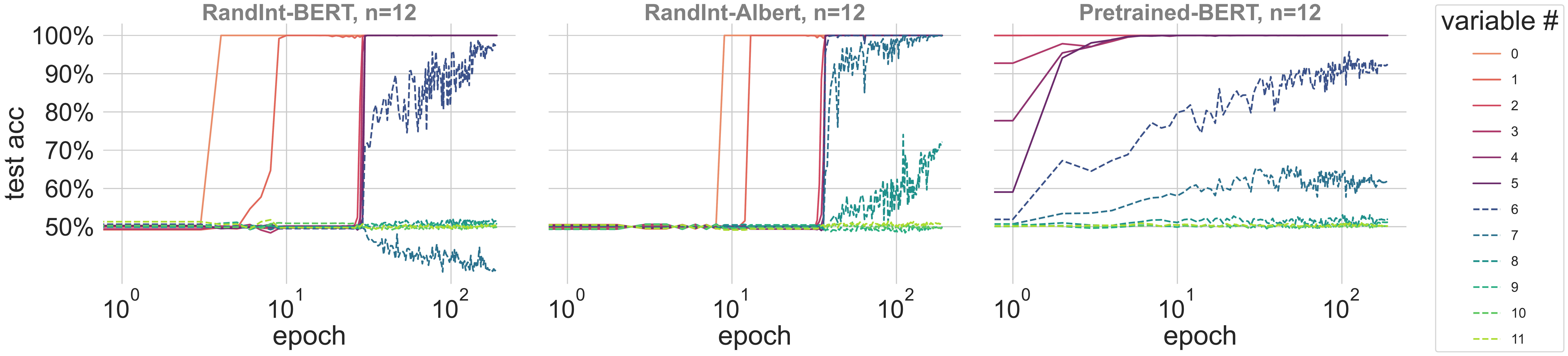}
        \end{minipage}
    \end{subfigure}
    \vspace{0.1cm}
    \hrule
    \vspace{0.1cm}
    \begin{subfigure}{\textwidth}
        \begin{minipage}[r]{0.1\textwidth}
        \caption{$n=16$}
        \end{minipage}
        \begin{minipage}[l]{0.85\textwidth}
            \centering
            \includegraphics[trim={0cm 0cm 0cm 0cm},clip, width=13cm]{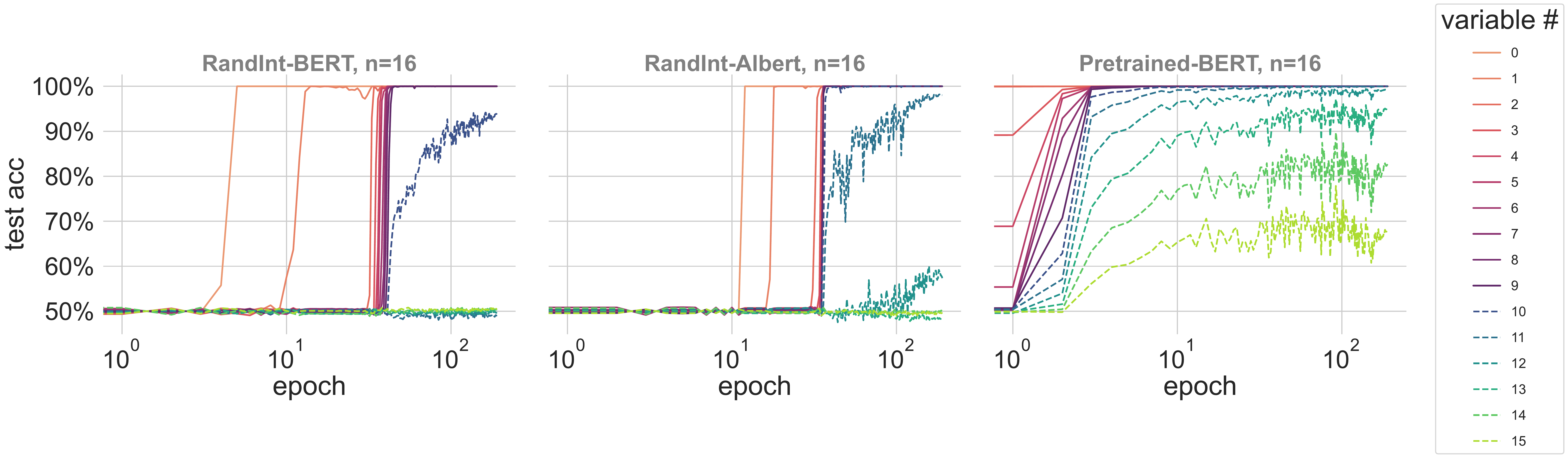}
        \end{minipage}
    \end{subfigure}
    \vspace{0.1cm}
    \hrule
    \vspace{0.1cm}
    \begin{subfigure}{\textwidth}
        \begin{minipage}[r]{0.1\textwidth}
        \caption{$n=20$}
        \end{minipage}
        \begin{minipage}[l]{0.85\textwidth}
            \centering
            \includegraphics[trim={0cm 0cm 0cm 0cm},clip, width=13cm]{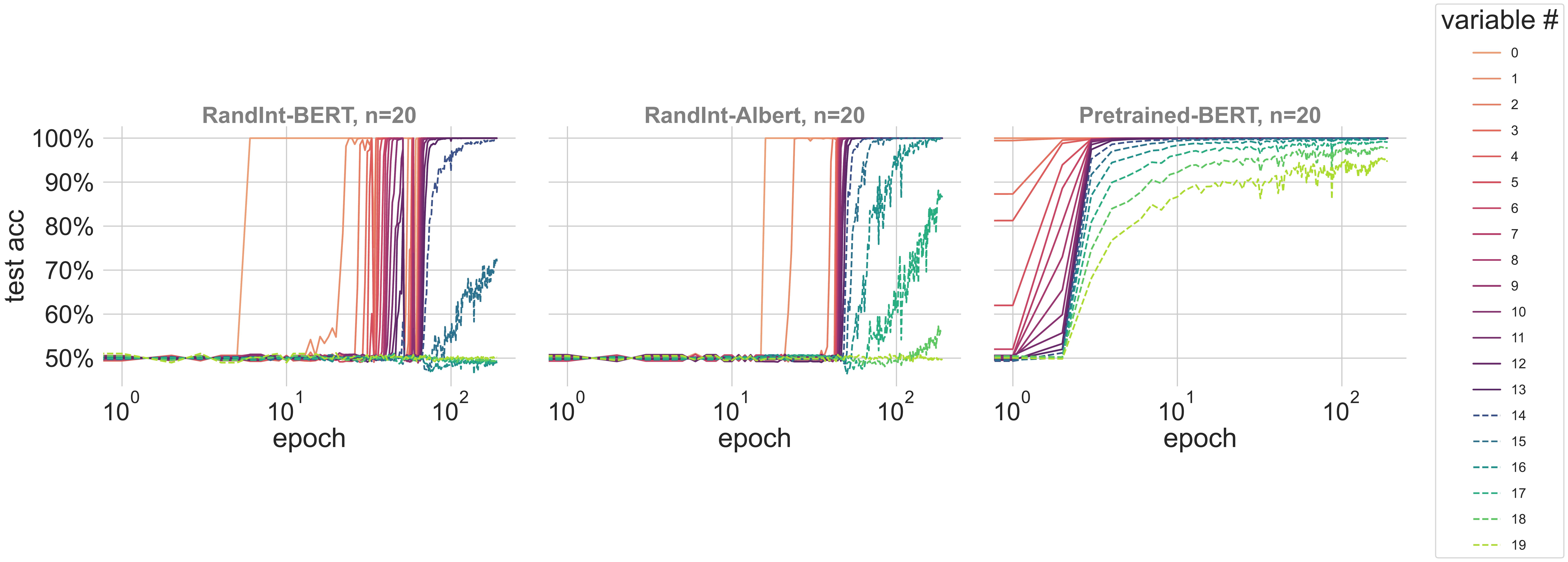}
        \end{minipage}
    \end{subfigure}
    \caption{Generalization performance of BERT and ALBERT models (depth $D=12$) trained on LEGO task of varying chain lengths $n$: (a) $n=8$, (b) $n=12$, (c) $n=16$, (d) $n=20$. }
    \label{fig:varying-n}
\end{figure}
We study this behavior by first repeating the experiments from our main paper on depth $D=12$ models on LEGO tasks of varying lengths $n$. In all the cases, we proportionally increase the length of chain for which supervision is provided by training on first $n_{tr}=n-6$ clauses in the chain.  

In Figure~\ref{fig:varying-n}, we show the performance of a typical run of transformer models in this setting. When trained on small length chains of $n=8$ and $n_{tr}=2$, we indeed see that  all the models including pretrained-BERT is able to learn the short training length (classical generalization) but the supervision does not contain enough information for the models to learn to generalize to unseen lengths. On the other hand, for larger chain lengths we see that the generalization only gets better, with a very strong monotonic trend for pretrained-BERT models. In particular, the strong generalization performance at $n=20$ with $n_{tr}=14$ would suggest that these models might not really be implementing the ``natural" iterative solution for the task. Rather, it is possible that training on longer sequences leads the models to  learn a more compact representation that nevertheless generalizes remarkably well to much longer lengths than seen during training.

\begin{figure}[H]
    \centering
    \begin{minipage}[l]{0.75\textwidth}
    \begin{subfigure}{\textwidth}
        \begin{minipage}[r]{0.2\textwidth}
        \caption{$D=2$}
        \end{minipage}
        \begin{minipage}[l]{0.75\textwidth}
            \centering
            \includegraphics[trim={0cm 0cm 0cm 0cm},clip, width=6cm]{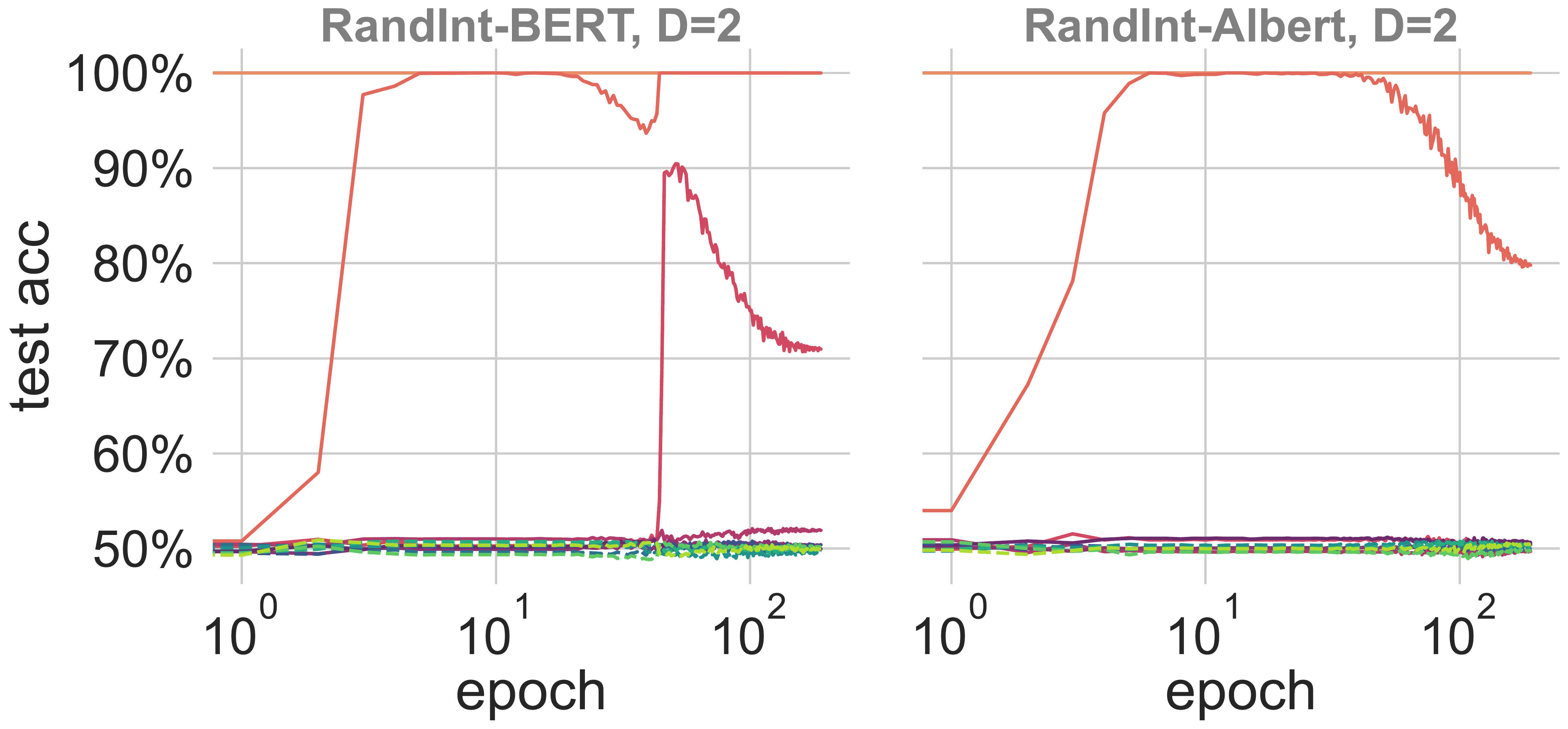}
        \end{minipage}
    \end{subfigure}

    \begin{subfigure}{\textwidth}
        \begin{minipage}[r]{0.2\textwidth}
        \caption{$D=4$}
        \end{minipage}
        \begin{minipage}[l]{0.75\textwidth}
            \centering
            \includegraphics[trim={0cm 0cm 0cm 0cm},clip, width=6cm]{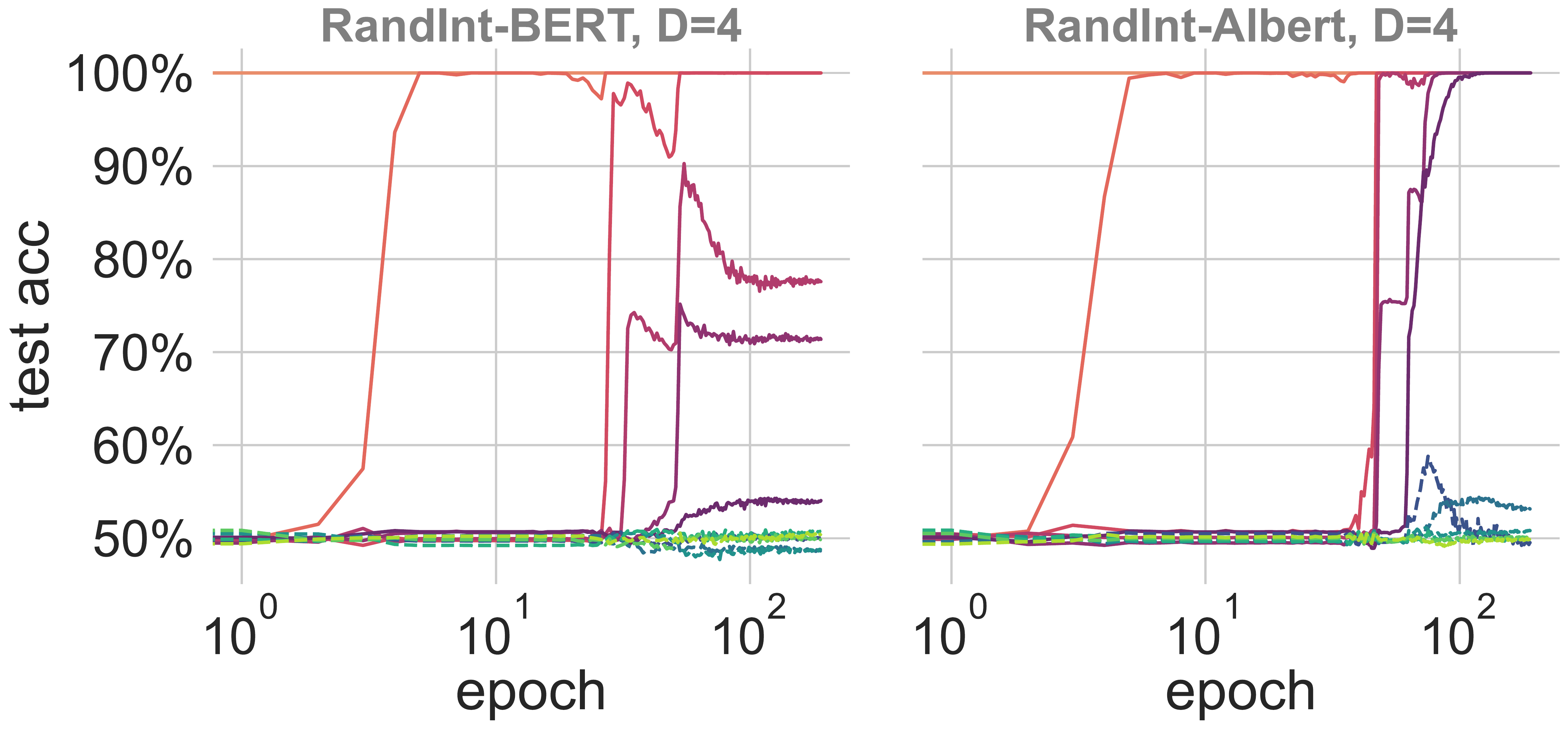}
        \end{minipage}
    \end{subfigure}

    \begin{subfigure}{\textwidth}
        \begin{minipage}[r]{0.2\textwidth}
        \caption{$D=8$}
        \end{minipage}
        \begin{minipage}[l]{0.75\textwidth}
            \centering
            \includegraphics[trim={0cm 0cm 0cm 0cm},clip, width=6cm]{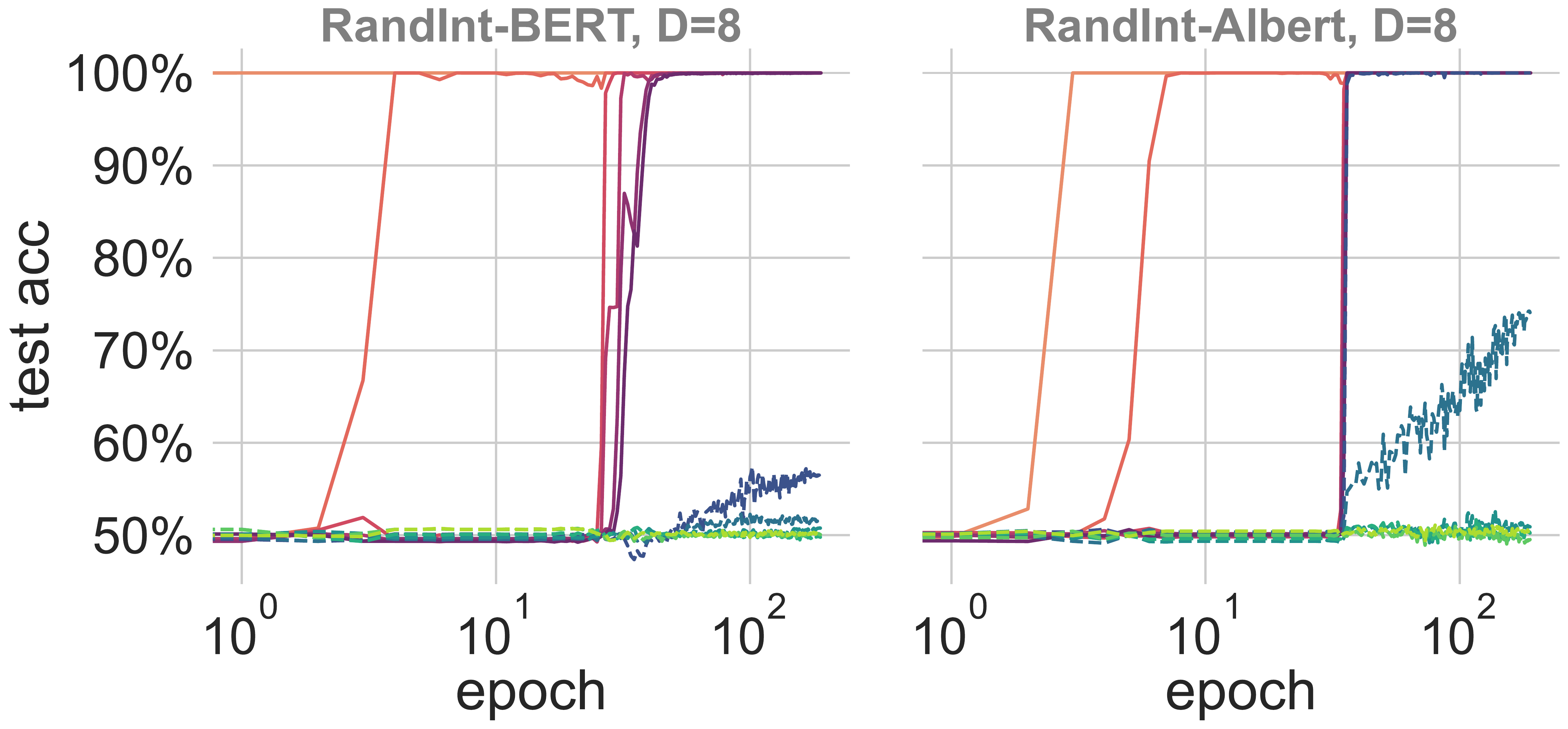}
        \end{minipage}
    \end{subfigure}
    \begin{subfigure}{\textwidth}
        \begin{minipage}[r]{0.2\textwidth}
        \caption{$D=12$}
        \end{minipage}
        \begin{minipage}[l]{0.75\textwidth}
            \centering
            \includegraphics[trim={0cm 0cm 0cm 0cm},clip, width=6cm]{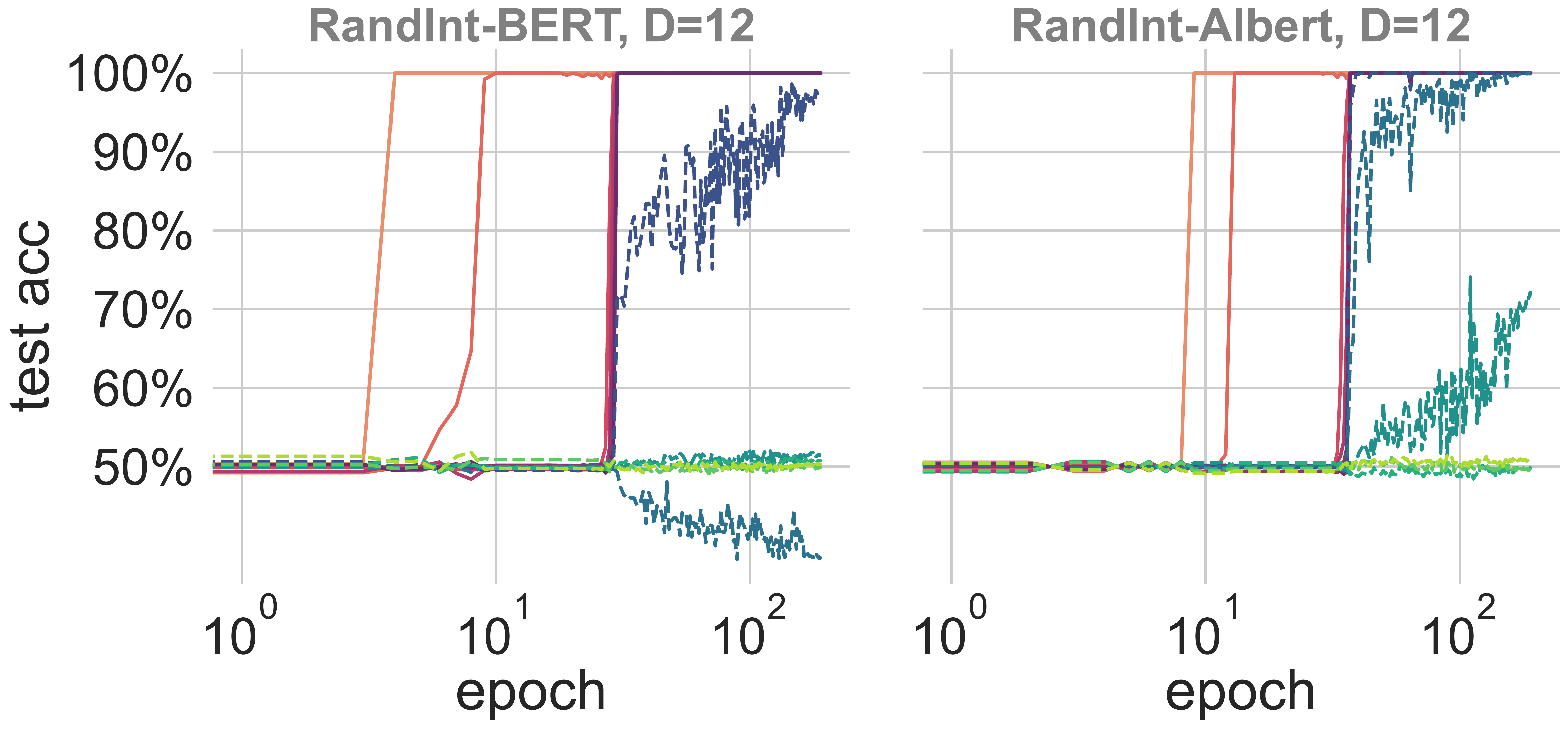}
        \end{minipage}
    \end{subfigure}
    \end{minipage}
    \begin{minipage}[l]{0.20\textwidth}
        \centering
        \includegraphics[trim={0cm 0cm 0cm 0cm},clip, width=1.2cm]{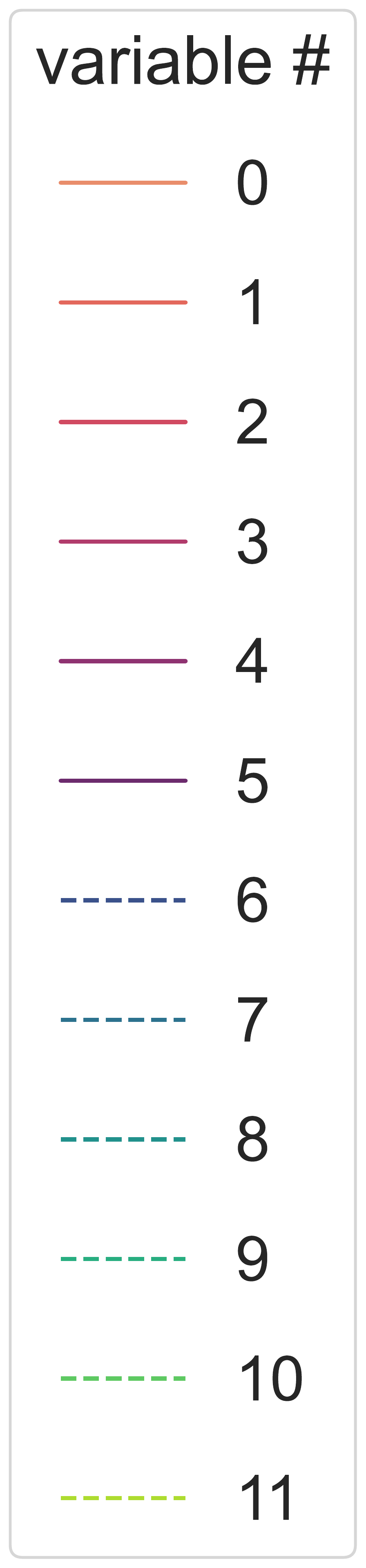}
    \end{minipage}
    
    \caption{Rand-Init BERT and ALBERT models of varying depth $D$ trained on LEGO tasks of length $n=12$.}
    \label{fig:varying-depth}
\end{figure}

Complementing our results on varying the length of LEGO chains, we also look at how  BERT and ALBERT architectures of smaller depth $D<12$ learn our LEGO task of length $n=12$. In Figure~\ref{fig:varying-depth} we show the results of models trained from random initialization (we do not have pre-trained models at these depths). Here we do see the trend that larger depth improves generalization. In particular, for a task of chain length $n$, there does appear to be a minimum threshold of $D$ at which the models learn to generalize even in the classical sense (on lengths the models were trained on), but this relationship appears sub-linear rather than linear with $D\ge n$ as speculated by the ``natural" iterative algorithm. It would be of interest in future studies to explore this relation between depth and length better. 

\subsection{Effect of number of parameters}
\begin{figure}[H]
\centering
\includegraphics[trim={0cm 0cm 0cm 0cm},clip, width=12cm]{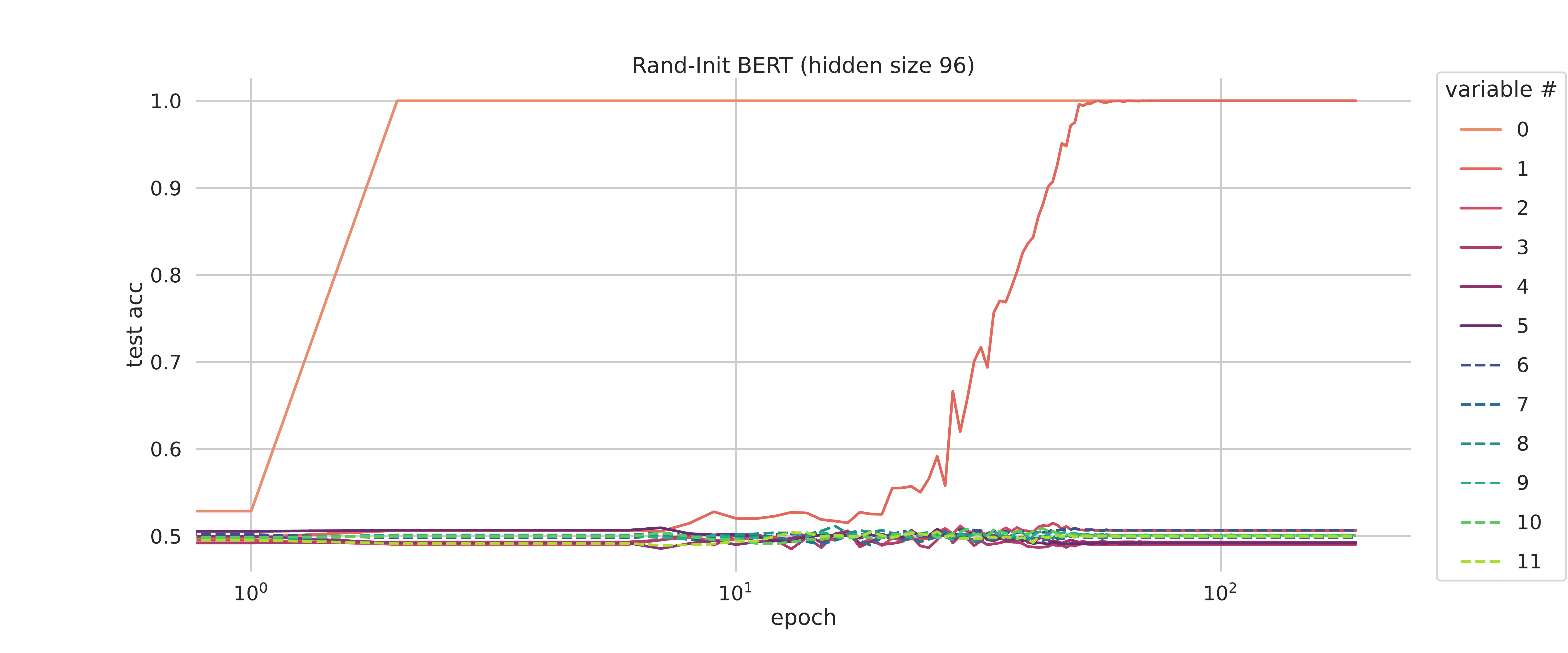}
\caption{The "thin" model not only lacks the length extrapolation properties of ALBERT, but it does not even generalize in-distribution. In fact, test accuracy on variable \#2 remains near 50\% after 200 training epochs. }
\label{fig:small-bert}
\end{figure}

In this section, we provide empirical evidence that the advantage of ALBERT over BERT for predicting longer sequences does not merely come from fewer parameters. Precisely, we modify the original BERT model to have 96 hidden size instead of 784 uniformly across all layers, with all other hyper-parameter unchanged. This "thin" BERT model has approximately the same parameter count as the ALBERT model. The result is shown in Figure~\ref{fig:small-bert}.

\section{LEGO Task 2: dihedral group}
\label{sec:dihedral}
As a generalization of  the main task (i.e., LEGO Task 1) analyzed so far, we present LEGO Task 2 of learning the dihedral group\footnote{\url{https://en.wikipedia.org/wiki/Dihedral_group}} $D_3$ of order $6$ which is isomorphic to the symmetric group $S_3$. Note that LEGO Task 1 can be viewed as learning the dihedral group $D_1$ of order $2$. 
Clearly, the shortcut solution described in Section~\ref{app:shortcut} is not valid here.


We repeat the out-of-distribution generalization experiments on Task 2 with the exact same model configurations and training hyper-parameters as Task 1 (see Section~\ref{training}). For dataset creation, we largely follow the pipeline detailed in Section~\ref{sec:data-generation} with group elements from $D_3$ for which we create corresponding tokens. The only modification here is that the labels are categorical with $6$ classes, since every variable in Task 2 may take $6$ candidate values.
\begin{figure}[htb] 
    \centering
    \includegraphics[trim={0cm 0cm 0cm 0cm},clip, width=13cm]{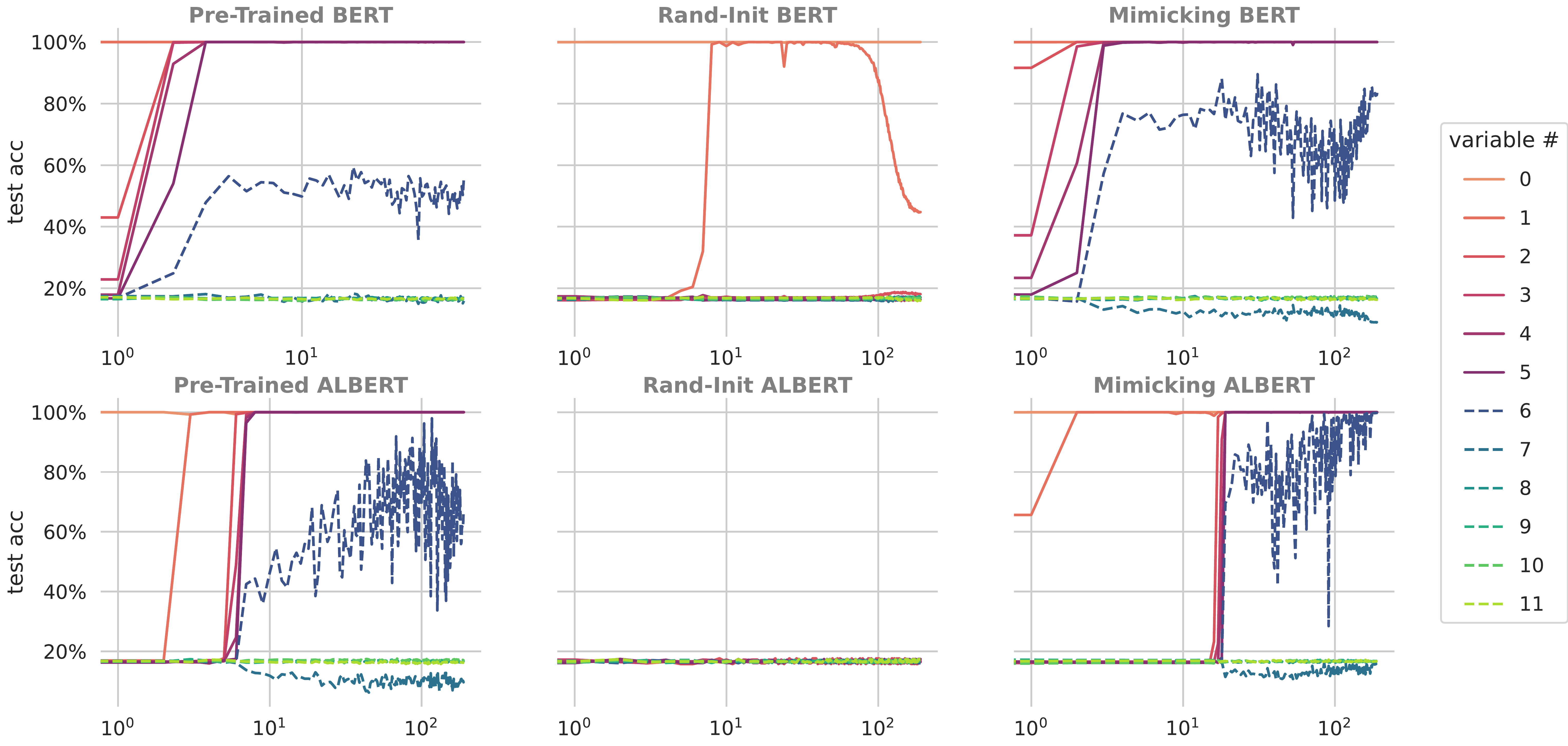}
    \caption{Out-of-distribution generalization on LEGO Task 2. Task 2 appears to be significantly more challenging than Task 1 and pretraining plays a more important role on top of rand-init models, while the mimicking procedure is able to match and even outperform pretraining.}
    \label{fig:dihedral}
\end{figure}

In Figure~\ref{fig:dihedral}, we show these preliminary results and observe that Task 2 is indeed more challenging than Task 1, and the extent to which pretraining provides benefits is noticeably larger. Without further hyper-parameter tuning, the randomly initialized models even face optimization issues in fitting the training labels. Interestingly, we also find our proposed mimicking procedure introduced in Section~\ref{sec:mimic} is not only able to match pretraining's performance and also outperform it. 

So far, none of the models here is capable of non-trivially generalizing to more than one extra variable. It is an important future direction to search for suitable adaptations to the current transformer architectures/training algorithms that can eventually solve this task. 

Furthermore, the amount of knowledge that models have to "memorize", e.g. the outcomes of $g(x)$ for all $g\in G$ and $x\in X$ grows quadratically with group size. While LEGO task 1 contains only 4 mappings, namely $1=+1, -1=+ (-1), 1 = - (-1), -1 = + (-1)$. For moderately size group such as $D_5$, models will have to "memorize" $(5!)^2=125^2$ in total mappings. This makes the task perfectly suitable for analyzing how and where (at which neurons) the models store these informations. We consider this as an exciting future direction.



\newpage
\section{Attention maps of pretrained BERT on LEGO}
\label{sec:attn-maps}
To complement the visualizations in Figure~\ref{fig:attention_heads}, we provide visualizations of attentions maps of all attention heads from a pretrained BERT when taking a LEGO sequence as input. We make the observation that \emph{almost every} attention head implements either a manipulation or an association operator. Please view electronically and zoom in for details. All the figures are vectorized.
\begin{figure}[H]
    \centering
    \begin{subfigure}{.5\textwidth}
            \centering
            \includegraphics[trim={0cm 0cm 0cm 0cm},clip, width=7cm]{figures/attention_maps/attention_layer_0.pdf}
            \caption*{layer 0}
    \end{subfigure}%
    \begin{subfigure}{.5\textwidth}
            \centering
            \includegraphics[trim={0cm 0cm 0cm 0cm},clip, width=7cm]{figures/attention_maps/attention_layer_1.pdf}
            \caption*{layer 1}
    \end{subfigure}
    \label{fig:all-attention}
\hfill
    \begin{subfigure}{.5\textwidth}
            \centering
            \includegraphics[trim={0cm 0cm 0cm 0cm},clip, width=7cm]{figures/attention_maps/attention_layer_2.pdf}
            \caption*{layer 2}
    \end{subfigure}%
    \begin{subfigure}{.5\textwidth}
            \centering
            \includegraphics[trim={0cm 0cm 0cm 0cm},clip, width=7cm]{figures/attention_maps/attention_layer_3.pdf}
            \caption*{layer 3}
    \end{subfigure}
    \label{fig:all-attention}
\hfill
    \begin{subfigure}{.5\textwidth}
            \centering
            \includegraphics[trim={0cm 0cm 0cm 0cm},clip, width=7cm]{figures/attention_maps/attention_layer_4.pdf}
            \caption*{layer 4}
    \end{subfigure}%
    \begin{subfigure}{.5\textwidth}
            \centering
            \includegraphics[trim={0cm 0cm 0cm 0cm},clip, width=7cm]{figures/attention_maps/attention_layer_5.pdf}
            \caption*{layer 5}
    \end{subfigure}
    \label{fig:all-attention}
\end{figure}
\vfill
\newpage

\newpage

\begin{figure}[H]
    \centering
    \begin{center} 
    \begin{subfigure}{.5\textwidth}
            \centering
            \includegraphics[trim={0cm 0cm 0cm 0cm},clip, width=7cm]{figures/attention_maps/attention_layer_6.pdf}
            \caption*{layer 6}
    \end{subfigure}%
    \begin{subfigure}{.5\textwidth}
            \centering
            \includegraphics[trim={0cm 0cm 0cm 0cm},clip, width=7cm]{figures/attention_maps/attention_layer_7.pdf}
            \caption*{layer 7}
    \end{subfigure}
    \label{fig:all-attention}
\hfill
    \begin{subfigure}{.5\textwidth}
            \centering
            \includegraphics[trim={0cm 0cm 0cm 0cm},clip, width=7cm]{figures/attention_maps/attention_layer_8.pdf}
            \caption*{layer 8}
    \end{subfigure}%
    \begin{subfigure}{.5\textwidth}
            \centering
            \includegraphics[trim={0cm 0cm 0cm 0cm},clip, width=7cm]{figures/attention_maps/attention_layer_9.pdf}
            \caption*{layer 9}
    \end{subfigure}
    \label{fig:all-attention}
\hfill
    \begin{subfigure}{.5\textwidth}
            \centering
            \includegraphics[trim={0cm 0cm 0cm 0cm},clip, width=7cm]{figures/attention_maps/attention_layer_10.pdf}
            \caption*{layer 10}
    \end{subfigure}%
    \begin{subfigure}{.5\textwidth}
            \centering
            \includegraphics[trim={0cm 0cm 0cm 0cm},clip, width=7cm]{figures/attention_maps/attention_layer_11.pdf}
            \caption*{layer 11}
    \end{subfigure}
    \label{fig:all-attention}
    \end{center}
\end{figure}
\makeatother

\end{document}